\documentclass[journal]{IEEEtran}
\usepackage{acro}
\newcommand{\newac}[2]{\DeclareAcronym{#1}{short=#1,long=#2}}

\newac{CNES}    {French National Centre for Space Studies}
\newac{CV}      {Computer Vision}
\newac{DEM}     {Digital Elevation Map}
\newac{ESA}     {European Space Agency}
\newac{ESTEC}   {European Space TEchnology and Research Centre}
\newac{FMM}     {Fast Marching Method}
\newac{FoV}     {field of view}
\newac{GA SLAM} {Globally Adaptive SLAM}
\newac{GNSS}    {Global Navigation Satellite System}
\newac{HDPR}    {Heavy Duty Planetary Rover}
\newac{HJB}     {Hamilton-Jacobi-Bellman}
\newac{ICP}     {Iterative Closest Point}
\newac{IMU}     {Inertial Measurement Unit}
\newac{LocCam}  {Localization Cameras}
\newac{MER}     {Mars Exploration Rovers}
\newac{MRO}     {Mars Reconnaissance Orbiter}
\newac{MSL}     {Mars Science Laboratory}
\newac{NavCam}  {Navigation Cameras}
\newac{NASA}    {National Aeronautics and Space Administration}
\newac{OUM}     {Ordered Upwind Method}
\newac{bi-OUM}   {bi-directional Ordered Upwind Method}
\newac{PRL}     {Planetary Robotics Lab}
\newac{PTU}     {Pan-and-Tilt Unit}
\newac{PanCam}  {Panoramic Cameras}
\newac{PDE}     {Partial Differential Equation}
\newac{RAMS}    {Reliability\, Availability\, Maintainability and Safety}
\newac{ROCC}    {Rover Operations Control Centre}
\newac{RRT}     {Rapidly Random Trees}
\newac{RTK}     {Real Time Kinematic}
\newac{RoI}     {region of interest}
\newac{SLAM}    {Simultaneous Localization and Mapping}
\newac{TGO}     {Trace Gas Orbiter}
\newac{TRL}     {Technology Readiness Level}
\newac{ToF}     {time-of-flight}
\newac{UGV}     {Unmanned Ground Vehicle}
\newac{VO}      {Visual Odometry}

\usepackage{graphicx}
\usepackage{setspace}
\usepackage{xcolor}
\usepackage{hyperref}
\usepackage{multirow}

\usepackage{graphbox}
\usepackage{float}
\usepackage{subfig}
\usepackage{soul}

\usepackage{amsfonts}

\usepackage[normalem]{ulem}

\usepackage{cite}

\ifCLASSINFOpdf
\else
\fi

\usepackage{amsmath}

\usepackage{algpseudocode}

\begin{document}
\title{Optimal Path Planning using CAMIS: a Continuous Anisotropic Model for Inclined Surfaces}

\author{J.~Ricardo~S\'anchez-Ib\'a\~nez,
        Carlos~J.~P\'erez-Del-Pulgar,~\IEEEmembership{Member,~IEEE},
        Javier~Ser\'on,
        and~Alfonso~Garc\'ia-Cerezo,~\IEEEmembership{Senior~Member,~IEEE}%
\thanks{This work was supported by the Andalusian regional government, Spain, under grant reference P18-RT-991 and the Spanish government under grant reference RTI2018-093421-B-I00.}%
\thanks{Authors are with the Space Robotics Laboratory, Department
of Systems Engineering and Automation, Universidad de M\'alaga, M\'alaga,
C/Ortiz Ramos s/n, 29071 Spain (e-mail: ricardosan@uma.es; carlosperez@uma.es).}%
\thanks{Manuscript received - -, -; revised - -, -.}}

\markboth{IEEE Transactions on Robotics,~Vol.~XX, No.~X, MMMM~202Y}%
{Shell \MakeLowercase{\textit{et al.}}: Bare Demo of IEEEtran.cls for IEEE Journals}

\maketitle

\begin{abstract}
The optimal traverse of irregular terrains made by ground mobile robots heavily depends on the adequacy of the cost models used to plan the path they follow. The criteria to define optimality may be based on minimizing energy consumption and/or preserving the robot stability. This entails the proper assessment of anisotropy to account for the robot driving on top of slopes with different directions. 
To fulfill this demand, this paper presents the Continuous Anisotropic Model for Inclined Surfaces, a cost model compatible with anisotropic path planners like the bi-directional Ordered Upwind Method. This model acknowledges how the orientation of the robot with respect to any slope determines its energetic cost, considering the action of gravity and terramechanic effects such as the slippage. Moreover, the proposed model can be tuned to define a trade-off between energy minimization and \textit{Roll} angle reduction. The results from two simulation tests demonstrate how, to find the optimal path in scenarios containing slopes, in certain situations the use of this model can be more advantageous than relying on isotropic cost functions. Finally, the outcome of a field experiment involving a skid-steering robot that drives on top of a real slope is also discussed. 
\end{abstract}


\IEEEpeerreviewmaketitle

\section{Introduction}
\label{sec-introduction}

\IEEEPARstart{M}{obile} robots demand more and more autonomy as they are necessary in increasingly complex tasks with limited or even nonexistent human supervision. In the case of emergency response scenarios, they can support the search of victims in unstructured environments \cite{delmerico2019current}. Moreover, they are used in the field of precision agriculture to harvest and/or fumigate more efficiently \cite{gonzalez2020unmanned}. Autonomy also plays a key point in planetary exploration, allowing rovers to traverse longer distances in extraterrestrial environments while raising the scientific return as result \cite{gao2017review}.
Nevertheless, the ability to navigate by themselves implies that they must be not only capable of properly perceiving their environment but also of driving through it in an efficient way. In most of applications, a mobile robot should recognize the environment and evaluate the terrain conditions taking into consideration its inherent features, i.e. kinematics and dynamics. Once evaluated, it should be able to generate and follow a trajectory to reach a target position.

This concern can be dealt with path planning algorithms. They fundamentally serve to calculate a path that connects the robot location to any other reachable destination. It is pertinent that this route, given any terrain and robot conditions, is optimal. This means there can not exist any other feasible path that, connecting the same two locations, requires less resources (e.g. energy and/or time) to follow it. Nevertheless, finding the optimal path is not a trivial issue and hence many approaches to achieve this have been introduced in the last decades.

\subsection{Approaches to Optimal Path Planning}
Existing path planning strategies fundamentally differ in two aspects:
the procedure to map the environment in a usable format and the process to later retrieve the path from it.
In the first case, the maps can be dynamically created by repeatedly adding nodes according to previous existing ones. By starting with only the initial location, this expansion resembles the growth of a tree, where new branches (samples) are added in an iterative way. This is the case of the \ac{RRT} algorithm, a so-called \textit{Sampling} algorithm that has been used for outdoor navigation \cite{krusi2017driving}. During the expansion, when one of those samples falls relatively close to the target point, a path can be traced backwards. This is done by looking for the parent node of each one in an iterative way until the original branch is reached. Moreover, this algorithm is asymptotically optimal: more branches can be created before retrieving the path to incrementally find better solutions as time runs. In this way, the algorithm can ultimately converge to the global optimum. Nevertheless, this entails high demand of memory resources, since usually large numbers of iterations and samples are needed to come relatively close to the global optimal solution \cite{noreen2016optimal}.

Contrary to this approach, a static structure made up in advance by interconnected nodes (a grid) serves as the starting point of other kinds of path planning algorithms. The quality of the resulting path depends on the procedure chosen to get it from such grid. We distinguish two groups: the \textit{Grid Search based} methods and the \textit{\ac{PDE} based} methods. Regarding the first group, they retrieve the path in a similar way as \ac{RRT}, making a reverse search. They differ in the manner the parenthood between nodes is established. The most basic of them is Dijkstra \cite{dijkstra1959note}, being used recently for guiding a rover in a simulated lunar scenario \cite{sutoh2015right}. Heuristic methods like A* evolved from Dijkstra to speed up computation \cite{hart1968formal}. A further improvement came in the form of incremental methods, which allowed to recycle previous computation to tackle maps with dynamic cost. However, all these algorithms suffer from the restriction of obtaining a path constrained to the topology of the grid, i.e. made up by consecutive edges. This is translated into getting paths that may not be smooth enough to be followed by a robot without unnecessary turns.
To overcome this restriction (and hence find better solutions), a subfamily of \textit{Grid Search based} methods arose: the \textit{Any-angle} algorithms. One of them is Field-D*, an algorithm implemented on the NASA rovers for the \ac{MER} and \ac{MSL} missions \cite{carsten2007global, maimone2007overview}.
It introduces the possibility of connecting neighbouring nodes by segments non contained within the grid edges, using interpolation methods in the process. Although this algorithm is incremental, it does not ensure to find the global optimal path. More recent strategies are focused on improving the quality of the resulting paths, heuristically looking for parent nodes beyond the node neighbourhood. This is the case of Theta*, which provides shorter paths than Field-D* while reducing the turns introduced by the latter \cite{nash2007theta}. However, it still can not ensure that the final solution is globally optimal \cite{daniel2010theta}. Many other improvements have been introduced to obtain even shorter paths and comparisons between them can be found in the literature \cite{uras2015speeding,harabor2016optimal}. Nevertheless, they are exclusively focused on maps distinguishing exclusively between obstacle and non-obstacle nodes, putting the use of non-uniform non-obstacle cost aside. Other works propose improvements on Theta* to make it compatible with multiple values of cost \cite{choi2011any}, but still direction changes can only occur on edges or nodes. A more recent algorithm called \textit{3DANA} arises as an \textit{Any-angle} method that makes use of 2.5D grids, i.e. 2D grids with elevation, to provide a path according to heuristic functions to control steering changes and avoid high magnitudes of slope \cite{munoz20173dana}.

Nevertheless, there is another group of path planning algorithms able to generate smooth paths in a globally optimal way: the \ac{PDE} based methods. Unlike \textit{Grid-Search based} methods, they do not compute any parent-child relation between nodes, avoiding the necessity to stick the path to intermediate nodes or edges. Instead, these algorithms assign a \textit{characteristic} direction to each node, which correspond to the optimal heading of any path that crosses it.
This direction comes from solving the \ac{PDE} in the respective node \cite{sethian2003ordered}. The form of this equation determines which solver methods are suitable. For example, the \ac{FMM} is a numerical method created to exclusively solve the \textit{Eikonal} equation on each node, visiting them in a similar fashion as in Dijkstra \cite{sethian1999fast}. The \textit{Eikonal} equation models the expansion of a wave that propagates with variable speed, coming in function of the existing cost values. This comes handy to find optimal paths in non-uniform cost maps \cite{liu2017predictive}. A recent survey provides further insight into more methods aimed at solving this \textit{Eikonal} equation, referring to them as \textit{Fast} methods \cite{gomez2019fast}.
In the case of the \textit{Eikonal}, these cost values are scalar and hence they do not depend on the wave direction. This also implies that starting the wave from either the initial or final point does not change the optimal solution. For this reason, the \textit{Eikonal} is categorized as isotropic. However, as will be seen later, some forms of cost require the consideration of the vehicle heading. This means the cost may change with direction, i.e. it may be anisotropic. 
There are efforts to adapt the \ac{FMM} to anisotropic cost under certain conditions due to its relatively cheap computational speed (O($\chi \log \chi$) where $\chi$ is the number of nodes). Some approaches consider cost directions parallel to the reference frame axes \cite{petres2005underwater,xu2019fast}, but for most unstructured scenarios this method is not suitable and hence produces sub-optimal results \cite{sethian2003ordered}. This can be overcome by using the \ac{OUM}, a numerical solver method that works using the static \ac{HJB} equation \cite{shum2016convergence}. In fact, the \textit{Eikonal} is the particular isotropic case of this equation, which admits the use of cost functions that vary with direction, i.e. anisotropic functions. Contrary to \ac{FMM}, \ac{OUM} can hence work with anisotropic cost functions, but it increases the computational load in exchange, proportionally to the existing anisotropy.
The main drawback of \ac{PDE} algorithms is the proper formatting of the cost map in order to make it continuous and differentiable, towards avoiding degenerated solutions in the form of non-smooth or the appearance of undesired zig-zag patterns in the path. 

\subsection{Anisotropic Models for Traversing Slopes}

To find optimal but also feasible paths on irregular terrains, it is key to properly assess the terrain-vehicle interaction. The traverse of uneven surfaces, for instance, is inherently anisotropic.  This is due to how in many ways the difference between the directions that the vehicle and the inclination are facing take relevance on the drive. One of these ways is, for example, how to look out for overturning. In the case of opting for using isotropic functions to model this (e.g. using a scalar weighted function only dependant on the magnitude of the slope), 
\ac{FMM} seems, at first, like a valid option to find optimal paths \cite{miro2010kyno}.
However, this approach may be indeed either too conservative or too permissive. This is because information regarding the differences in the longitudinal and lateral axes directions is lost, which may be instead preserved by means of an anisotropic function compatible with \ac{OUM} \cite{shum2015direction}. Apart from this risk assessment, there are more physical phenomenons taking effect in the traverse of inclined surfaces that depend on direction. For instance, the slip (ratio between the commanded and real vehicle speeds) is one of them. There is a recent research oriented to find paths for the optimal ascension of slopes using \ac{RRT}* \cite{inotsume2020robust}, where the difference in slip between going straightly or diagonally through them is considered \cite{inotsume2016finding}. Anisotropic functions to model power consumption according to friction and gravity effects were also used in the past along with \textit{Grid-Search} planners, such as modified versions of Dijkstra \cite{sun2005finding} and other heuristic functions \cite{rowe1990optimal,choi2012global,ganganath2015constraint,ganganath2018shortest}. It is worth mentioning that while not ensuring finding the globally optimal path, these \textit{Grid-Search} approaches allow the use of heavy discontinuities such as forbidding ascending certain inclinations or substituting them by discontinuous zig-zag maneuvers \cite{ganganath2015constraint}.

\subsection{Continuous Optimal Anisotropic Model}
 
The usage of an anisotropic cost model combining slip, friction and risk metrics for path planning purposes 
arises as an interesting research topic.
The closest approach found is a planner that produces bezier-curves as trajectories for ascending through inclined planes, as result of tuning some terramechanic parameters in a neural network model \cite{sakayori2017energy}. Besides, as denoted we have not found in the literature anisotropic \ac{PDE} planners working with these parameters even by separate. 

Therefore, the main purpose of this paper is to fill this gap towards finding the optimal path for a mobile robot in irregular terrains containing slopes. To do this, we present in this paper two contributions: first, the use of a cost model for path planning compatible with \ac{OUM}, presenting anisotropy (value dependant on direction), continuity (fully defined in all directions) and smoothness (continuously differentiable in all direction). We name this model \textit{CAMIS}, which is the acronym for \textit{Continuous Anisotropic Model for Inclined Surfaces}. Second, the use of \textit{CAMIS} to encompass multiple dynamic behaviours regarding the navigation through slopes, such as the effect of gravity, the friction, the slip and the risk to overturn. As result, we pretend to generate the optimal path, which is the one that provides the best trade-off between minimizing energy and preserving safety in scenarios containing uneven terrains.
Ultimately, we support these contributions by carrying out both simulation and field tests, where we thoroughly evaluate the versatility of using \textit{CAMIS}.

The structure of this paper comes in the following form. Section \ref{sec-problemstatement} gives an overview of the optimization problem involved in continuous path planning and its application on irregular terrains, focusing on the algorithm functioning and the cost function requirements. Later on, Section \ref{sec-camis} fully describes \textit{CAMIS}, starting from its mathematical background and then being followed by an explanation of how it considers friction, slip and several risks. Next, the results from a series of simulation tests using \textit{CAMIS} are introduced in Section \ref{sec-testing}, which serve to validate its potential. Last, Section \ref{sec-conclusions} provides our conclusions regarding the usage of \textit{CAMIS}, together with an overview of possible future work.

\section{Optimal Anisotropic Path Planning}
\label{sec-problemstatement}

Finding the optimal path to reach a certain destination in rough environments is not a trivial issue. First of all, we must clarify in which sense a path is \textit{optimal}. Then, the approach taken to figure out the shape of this path, i.e. the planning algorithm, is detailed in this section. Finally, we explain the main features of 
the \textit{CAMIS}, that must serve both as a compatible cost function to such algorithm and also as a tool to model the locomotion capabilities of any mobile robot on slopes.
\begin{figure}
    \centering
    \includegraphics[width=\columnwidth,align=c]{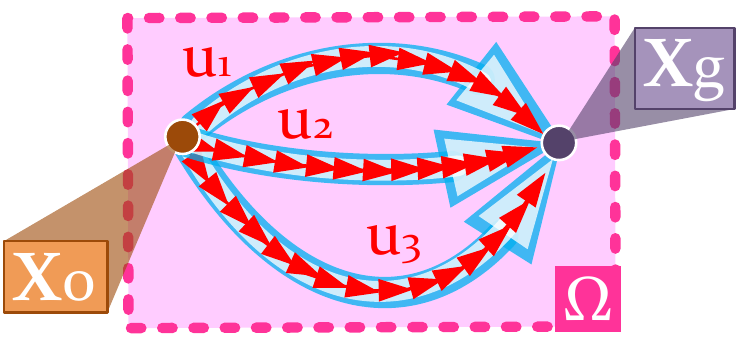}
    \caption{Different heading functions $u_1$, $u_2$ and $u_3$ produce paths (in blue) going from $x_0$ to $x_g$ within the $\Omega$ region.}
    \label{fig:problemstatement}
\end{figure}
\subsection{Problem Statement}
The objective path consists of a continuous curve located inside the closed region $\Omega \subset \mathbb{R}^2$.
We refer to it as the continuous and differentiable function $\Gamma : s \in \mathbb{R}_{+,0} \to x \in \Omega$, 
where $s$ is the distance covered from the origin position ($s=0$) to any point within it, whose maximum length is $s_g$. The function returns the position $x$ according to the covered distance $s$ within $\Omega$. Thereafter, as the path planning problem concerns anisotropy, i.e. dependency on direction, we define the heading function $u(s)$. It is the function that returns the unit tangent vector to $\Gamma(s)$ as denoted in 
(\ref{eq:pathDirection}).
\begin{equation}
\label{eq:pathDirection}
    u(s) = \frac{d \Gamma(s)}{ds}
\end{equation}
There are infinite ways to define a path $\Gamma$ connecting two whatever points $x_0 = \Gamma(s = 0)$ (origin) and $x_g = \Gamma(s = s_g)$ (goal). Hence, according to (\ref{eq:pathDirection}) there exists a set of infinite heading functions $U(x_0, x_g) = \left\{ u_1, u_2, ... u_\infty \right\}$ that could produce the aforementioned paths as result, as depicted in figure \ref{fig:problemstatement}. However, any of those paths entails a certain accumulation of cost going from $x_0$ to $x_g$, which is referred to here as \textit{Total Cost} or $T(x_0, x_g)$. Therefore, we formulate the path planning problem as finding the optimal heading function $u_i \in U(x_0, x_g)$ that entails the minimal value of $T(x_0, x_g)$. This is analogous to the approaches of many past research works \cite{sethian2003ordered, shum2015direction, shum2016convergence} based on the ideas of \textit{Dynamic Programming}, and results on  (\ref{eq:pathWaypoints}).
\begin{equation}
\label{eq:pathWaypoints}
    T(x_0, x_g) = \min_{u_i \in U(x_0, x_g)} \left\{ \int_{0}^{s_g} Q(\Gamma(s), u_i(s)) ds \right\}
\end{equation}
The anisotropy present in this problem comes from the \textit{Cost} function $Q$, defined by position and direction. This function provides positive nonzero cost values that are accumulated along the path. As a side note, we take the assumption that maybe the initial and final headings, $u(s=0)$ and $u(s=s_g)$ respectively, may not correspond to the initial and final desired headings of the robot. This is translated into the need of extra maneuvers to align the vehicle at the beginning and at the end of the traverse. This is no problem for those vehicles capable to turn in the spot (in exchange of a certain non considered amount of time/energy), while it is also an acceptable assumption for planning big traverses.
Moreover, the way $Q$ is defined means the cost derived from turning, i.e. changing heading, is omitted and not in the scope of this problem formulation. 
\begin{figure*}[t]
    \centering
    \includegraphics[width=\textwidth,align=c]{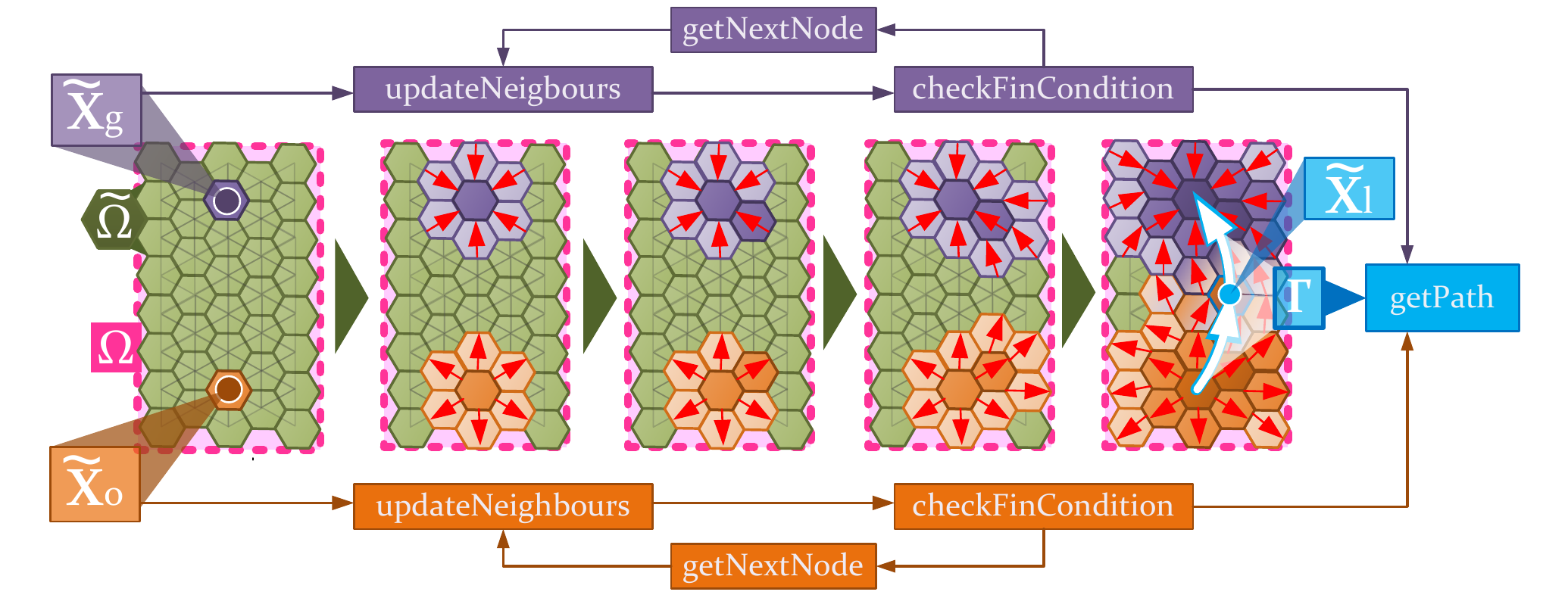}
    \caption{Two loops begin computing from the goal (in purple) and the start (in orange). Each loop iterate taking new nodes and computing two parameters: the \textit{Total Cost} and the characteristic direction passing through them (marked as red arrows).}
    \label{fig-biOUMprocess}
\end{figure*}
Equation (\ref{eq:pathWaypoints}) can be reformulated as the static \textit{Hamilton-Jacobi-Bellman} equation shown in (\ref{eq:HJB}). Here, for any position $x$ a certain unit vector of $\psi(x)$ is appointed. This vector corresponds to the so called \textit{characteristic direction}, which is the optimal heading of the path passing through $x$ and complying with (\ref{eq:HJB}), meaning $\psi(x) = u(s) \ | \ x = \Gamma(s)$.
\begin{equation}
  \label{eq:HJB}
    \min_{\psi(x)} \{ \psi(x) \nabla T(x_0,x) + Q(x, \psi(x)) \} = 0, \ \ \forall x \in \Omega
\end{equation}

By solving (\ref{eq:HJB}), not only we obtain $T(x)$ as result, but also $\psi(x)$.
To numerically solve this problem, as introduced in Section \ref{sec-introduction} there is already a fast method capable of achieving this: the \ac{OUM}. This algorithm produces a viscous solution of (\ref{eq:HJB}), in the sense it overcomes any discontinuity that may appear in the real solution of $T(x)$ \cite{shum2016convergence}. Furthermore, in order to preserve the uniqueness of $\psi(x)$, it is mandatory to make the speed profile function $F$, shown in (\ref{eq:propagation}), convex \cite{sethian2003ordered, shum2016convergence}. 
\begin{equation}
    \label{eq:propagation}
    F(x,\psi(x)) = \frac{1}{Q(x,\psi(x))}
\end{equation}

\begin{figure}
    \centering
    \includegraphics[width=\columnwidth,align=c]{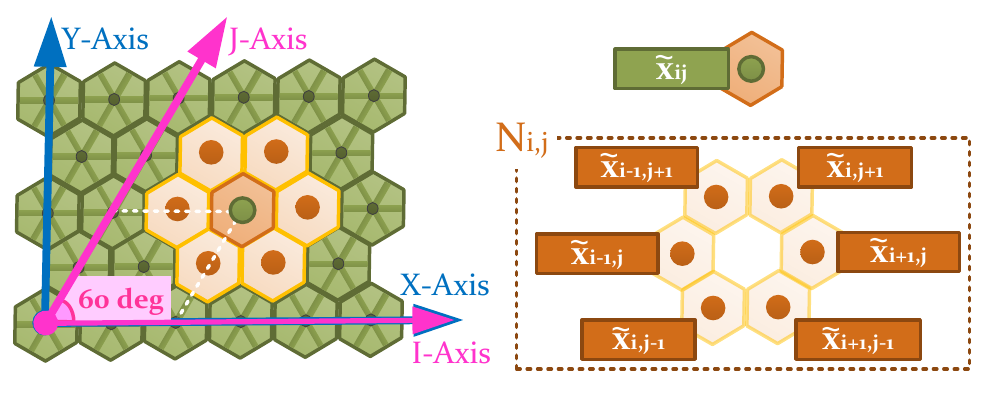}
    \caption{Nodes indexation, arranged as a regular hexagonal grid.}
    \label{fig-hexagonalAccess}
\end{figure}

\subsection{Path planning algorithm implementation}
As already mentioned, \ac{OUM} arises as a method capable to numerically solve equation \ref{eq:HJB} on a discretized $\Omega$, i.e. using a grid $\tilde{\Omega}$ to map the environment. This grid is made up by nodes $\tilde{x}$ that cover the entire region. \ac{OUM} accesses each of them to compute their corresponding values of $T$ and $\psi$. It can indifferently start from the origin position, computing hence $T(\tilde{x}_0, \tilde{x})$, or the goal, computing in this case $T(\tilde{x}, \tilde{x}_g)$, to later access the rest in a loop process. By exploiting this fact, an improved version named \ac{bi-OUM} allows to compute in parallel both of these options, stopping the loops when they reach an intermediate node $\tilde{x}_l$. This results on achieving the optimal solution as well but in a faster way as demonstrated in recent contributions \cite{shum2015direction}.

Figure \ref{fig-biOUMprocess} portrays a conceptual scheme of the functioning of \ac{bi-OUM}. We start by defining the already introduced grid $\tilde{\Omega}$, from which two nodes, $\tilde{x}_o$ and $\tilde{x}_g$, are selected. Thereafter, two propagating processes start from them, accessing neighbouring nodes in an iterative way. The access  to each node is done using discrete coordinates $(i,j)$ and a constant resolution $h$. For this particular case, we define the location of each node using (\ref{eq:discreteX}).
\begin{equation}
    \label{eq:discreteX}
    \tilde{x}_{ij} = [h i + h/2 j, \sqrt{3}/2 j ]
\end{equation} 
In this way, instead of relying on the use of an irregular grid we opt for a regular one. The main reason is because it can be easily stored and the indexation of each node can be sorted as denoted in (\ref{eq:discreteX}). Figure \ref{fig-hexagonalAccess} portrays this sorting, where for a node at $\tilde{x}_{ij}$, there are other six nearby forming the hexagon-shaped neighbourhood $N_{ij}$ presented in (\ref{eq:neighbourhood}).
\begin{equation}
\label{eq:neighbourhood}
    N_{ij} = \{\tilde{x}_{i+1j}, \tilde{x}_{ij+1}, \tilde{x}_{i-1j+1}, \tilde{x}_{i-1j}, \tilde{x}_{ij-1}, \tilde{x}_{i+1j-1}\}
\end{equation}

As a side note, the use of hexagons, although less common in robotics navigation, is for some cases more advantageous than squares \cite{algfoor2015comprehensive}. Further research about these advantages are found in applications such as computing the directional flow of a DEM to obtain an hydrological model \cite{de2006assessing} or for raster graphics \cite{brimkov2001honeycomb}. Another contribution\cite{de2006assessing}, shows how using this 6-neighbourhood scheme, instead of a regular square grid, demonstrates that less samples are required to represent a DEM while maintaining the resolution. Besides, the same authors point to the fact that an hexagon is closer to a circle than a square. Furthermore, all neighbours from any node are at the same distance $h$ to its center, i.e. isotropic distances, which makes hexagonal grids more symmetric and compact \cite{wang2020isotropic}.
\begin{figure}
\begin{algorithmic}[1]
\Function{biOUM}{$\tilde{x}_0,\tilde{x}_g, \tilde{\Omega}$}
\State \textbf{Input:} $\tilde{x}_0,\tilde{x}_g, \tilde{\Omega}$  
\State \textbf{Output:} $\Gamma$
\State $S_g(\tilde{x})  \gets (\infty, NaN, Far) \ \ \ \forall \tilde{x} \in \tilde{\Omega}$
\State $S_g(\tilde{x}_g) \gets (0, NaN, AcceptedFront)$
\State $S_0(\tilde{x})  \gets (\infty, NaN, Far) \ \ \ \forall \tilde{x} \in \tilde{\Omega}$
\State $S_0(\tilde{x}_0) \gets (0, NaN, AcceptedFront)$
\State $\tilde{x}_{tg} \gets \tilde{x}_g$
\State $\tilde{x}_{t0} \gets \tilde{x}_0$
\While{$\lnot checkFinCondition(S_g(\tilde{x}_{tg}), S_0(\tilde{x}_{tg})) \land$ \newline $\lnot checkFinCondition(S_g(\tilde{x}_{t0}), S_0(\tilde{x}_{t0}))$}
\State \Comment{Update both loops}
\State $updateNeighbours(\tilde{x}_{tg})$
\State $updateNeighbours(\tilde{x}_{t0})$
\State \Comment{Get next nodes}
\State $\tilde{x}_{tg} \gets getNextNode(S_g)$
\State $\tilde{x}_{t0} \gets getNextNode(S_0)$
\EndWhile
\State \textbf{return} $getPath(S_g, S_0)$
\EndFunction
\end{algorithmic}
\caption{The biOUM algorithm.}\label{fig-algorithm}
\end{figure}

Whenever each node $\tilde{x}$ is evaluated, the resulting information is stored in two sets $S_0$ and $S_g$. These sets are made up of tuples $S_0(\tilde{x}) = (T(\tilde{x}_0, \tilde{x}), \psi(\tilde{x}), state)$ and $S_g(\tilde{x}) = (T(\tilde{x}, \tilde{x}_g), \psi(\tilde{x}), state)$, where $state$ is a label that indicates the execution state of that node with respect to the corresponding loop. This state, can be either \textit{Far} (nodes are yet to be accessed), \textit{Considered} (assigned values are tentative), \textit{AcceptedFront} (assigned values are definitive but those of one or more neighbours not) and \textit{Accepted Inner} (assigned values of that node and neighbours are definitive). Therefore, the \textit{state} of both tuples of every node are initialized to \textit{Far}. 
Following the flowcharts depicted in figures \ref{fig-biOUMprocess} and \ref{fig-algorithm}, each loop sets the $Total Cost$ of the respective starting node to zero, a null value to $\psi$ (since initial and final desired orientations are omitted as mentioned before) and the \textit{Accepted} state. Thereafter, each process updates the neighbours of the corresponding starting node thanks to the \textit{updateNeighbours()} function. In an iterative way, the next node whose neighbours are meant to be updated is selected by using the \textit{getNextNode()} function. On the one hand, \textit{updateNeighbours()} consists of three parts: getting new \textit{AcceptedInner} nodes from \textit{AcceptedFront} nodes, getting new \textit{Considered} nodes from \textit{Far} nodes and updating the values of \textit{Total Cost} and the \textit{characteristic direction} of the latter ones. On the other hand, \textit{getNextNode()} takes the \textit{Considered} node with the lowest value of \textit{Total Cost} and labels it as \textit{AcceptedFront}.

\begin{figure}
    \centering
    \includegraphics[width=\columnwidth,align=c]{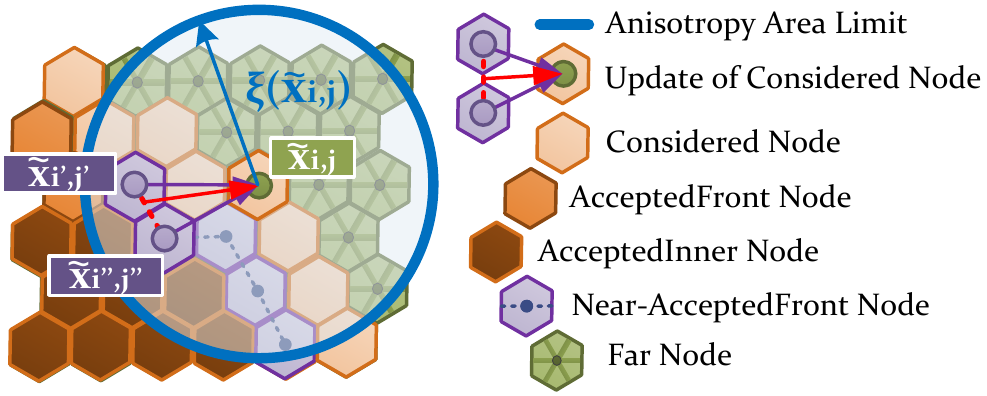}
    \caption{\textit{Total Cost} update of a considered node, using those \textit{AcceptedFront} nodes within the anisotropy area.}
    \label{fig-Tupdate}
\end{figure}

Regarding the process to update $T$ and $\psi$ for a \textit{Considered} node $\tilde{x}$, figure \ref{fig-Tupdate} shows a conceptual image of its functioning. It essentially depends on the closest \textit{AcceptedFront} nodes that are placed under a certain distance $\xi(\tilde{x})$ from the \textit{Considered} node in question. This distance, as seen in equation (\ref{eq:anisotropicdistance}), is proportional to not only the grid resolution $h$ but also to the anisotropy function $\Upsilon(\tilde{x})$. The latter function returns the ratio between the maximum and the minimum possible cost values according to $\psi$.
\begin{equation}
    \label{eq:anisotropicdistance}
    \xi(\tilde{x}_{ij}) = h \Upsilon(\tilde{x}_{ij})
\end{equation}
\begin{equation}
    \label{eq:anisotropydefinition}
    \Upsilon(\tilde{x}_{ij}) = \frac{\max\limits_{\psi} \{ Q(\tilde{x}_{ij},\psi) \} }{\min\limits_{\psi} \{ Q(\tilde{x}_{ij},\psi) \} }
\end{equation}

The update of 
$T(\tilde{x}_0,\tilde{x}_{ij})$ comes as in (\ref{eq:updateT}) using two types of discretization proposed in \cite{sethian2003ordered}: the semi-lagrangian and the eulerian. 
\begin{equation}
\label{eq:updateT}
    T(\tilde{x}_0,\tilde{x}_{ij}) = \begin{cases}
\min\limits_{\epsilon \in [0,1]} \{C(\tilde{x}_{ij},\psi(\tilde{x}_{ij})) |\epsilon \tilde{x}_{i'j'} \ + \\ \hfill \hfill 
(1 - \epsilon) \tilde{x}_{i''j''} - \tilde{x}_{ij}| 
+ \epsilon T(\tilde{x}_0,\tilde{x}_{i'j'}) \ + \\
\hfill \hfill (1-\epsilon) T(\tilde{x}_0,\tilde{x}_{i''j''}) \} \ , \\ 
\hfill \hfill \Upsilon(\tilde{x}_{ij}) > 1 \\ \\
\frac{T(\tilde{x}_0, \tilde{x}_{i'j'}) + T(\tilde{x}_0,\tilde{x}_{i''j''})}{2} + \frac{1}{2} ((T(\tilde{x}_0,\tilde{x}_{i'j'}) \ + \\ \hfill \hfill  T(\tilde{x}_0,\tilde{x}_{i''j''}))^2 +
3 (h C(\tilde{x}_{ij},\cdot))^2 \ + \\ \hfill \hfill  4 \left(T(\tilde{x}_0,\tilde{x}_{i'j'}) - T(\tilde{x}_0,\tilde{x}_{i''j''})\right)^2)^{1/2},\\ \hfill \hfill \Upsilon(\tilde{x}_{ij}) = 1 \ \land Eq.(\ref{eq:upwindCondition}) \\ \\  
\min \left\lbrace T(\tilde{x}_0,x_{i'j'}), T(\tilde{x}_0,x_{i''j''})\right\rbrace + \\ \hfill \hfill h C(x_{ij},\cdot)  \ , \\ \hfil \hfill \hfill otherwise
\end{cases}
\end{equation}
As stated before, this update is done using two neighbours labeled as \textit{AcceptedFront} and under a distance $\xi(\tilde{x}_{ij})$. 
We refer to these two neighbours $\tilde{x}_{i'j'}$
 and $\tilde{x}_{i''j''}$, as shown in figure \ref{fig-Tupdate}.
In the anisotropic case ($\Upsilon(\tilde{x}_{ij}) > 1$), the value of $T(\tilde{x}_0,\tilde{x}_{ij})$ is obtained using the semi-lagrangian discretization, but it is needed to iteratively approximate this value until it converges to the solution.
Contrary to this, we have added an explicit form to get $T$ in the isotropic case ($\Upsilon = 1$) for the regular hexagonal grid used here. This is obtained from the upwind finite-difference discretization proposed by \cite{sethian2003ordered}. 
In this way, terrains presenting a high degree of isotropy can benefit from this computationally cheap update method. 
Another way for the isotropic case is added, in those situations where the upwind condition (\ref{eq:upwindCondition}) cannot be fulfilled with the eulerian discretization. Then, the corresponding value of $\psi(\tilde{x}_{ij})$ is also computed following (\ref{eq:updateU}). 

It is important to mention that these two update processes in the case of \ac{bi-OUM} have a different treatment according to the loop they correspond. If they are executed within the loop starting from the goal node,
the cost function $Q$ is \textit{flipped}, in the sense that the expansion of this loop is contrary to the direction of the vehicle. For this reason, values of $\psi$ are the ones the vehicle must follow adding $\pi$ rads.
\begin{equation}
\label{eq:upwindCondition}
    T(\tilde{x}_0,\tilde{x}_{ij}) > \min \left\lbrace T(\tilde{x}_0,\tilde{x}_{i'j'}), T(\tilde{x}_0,\tilde{x}_{i''j''})\right\rbrace
\end{equation}
\begin{equation}
    \label{eq:updateU}
\psi(\tilde{x}_{ij}) = \begin{cases}
\epsilon \tilde{x}_{i'j'} + (1 - \epsilon) \tilde{x}_{i''j''} - \tilde{x}_{ij} , \\ \hfill \hfill \hfill \Upsilon(\tilde{x}_{ij}) > 1 \\ \\
\left[ \begin{array}{c} \frac{\tilde{x}_{ij} - \tilde{x}_{i'j'}}{h}\\ \frac{\tilde{x}_{ij} - \tilde{x}_{i'j'}}{h} \end{array} \right]^{-1} \left[ \begin{array}{c} \frac{T(\tilde{x}_0, \tilde{x}_{ij}) - T(\tilde{x}_0, \tilde{x}_{i'j'})}{h} \\ \frac{T(\tilde{x}_0, \tilde{x}_{ij}) - T(\tilde{x}_0, \tilde{x}_{i''j''})}{h} \end{array} \right], \\ \hfill \hfill \hfill otherwise
\end{cases}
\end{equation}

Finally, the \textit{checkFinCondition()} function determines whether the last node obtained with \textit{getNextNode()} is either \textit{AcceptedInner} or \textit{AcceptedFront} for both sets $S_g$ and $S_0$. The node complying with this condition, marked as a blue dot in figure \ref{fig-biOUMprocess}, is named $\tilde{x}_l$. From its location, two portions of path are extracted using the \textit{getPath()} function. This is thanks to successively computing \textit{waypoints}, i.e. discrete position samples, by following the computed $\psi$ values (abbreviated from $\psi(\tilde{x})$) until $\tilde{x}_0$ and $\tilde{x}_g$ are reached.
Then, the merge of both portions results on the final complete path $\Gamma$.

\section{Formulation of CAMIS}
\label{sec-camis}

\begin{figure}
    \centering
    \subfloat[3D Perspective view.
        \label{fig-slopeIso}]%
    	{\includegraphics[width=.6\columnwidth,align=c]{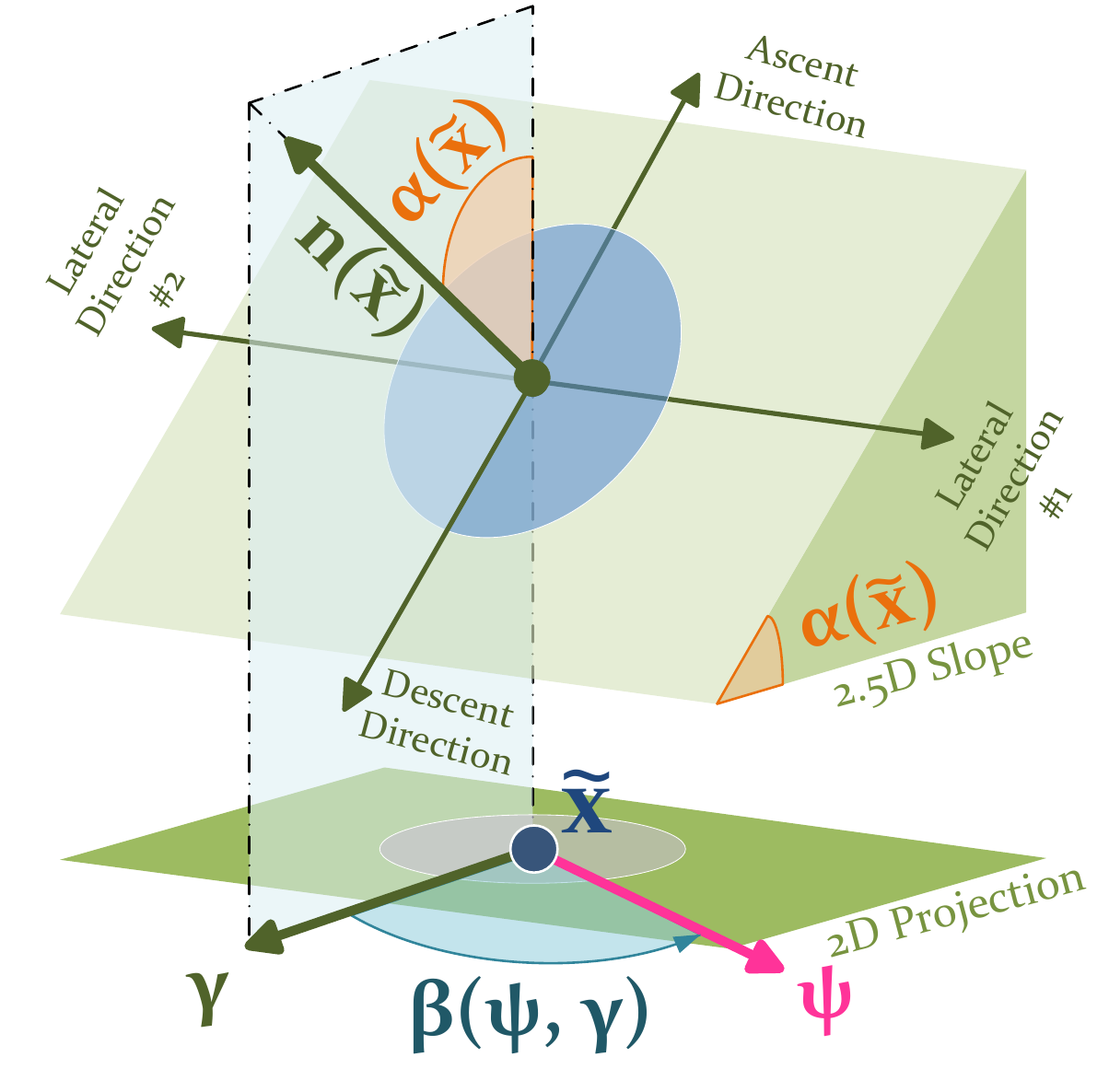}}
    \quad
    \subfloat[Lateral and top view.
        \label{fig-slopeAbove}]%
    	{\includegraphics[width=.34\columnwidth,align=c]{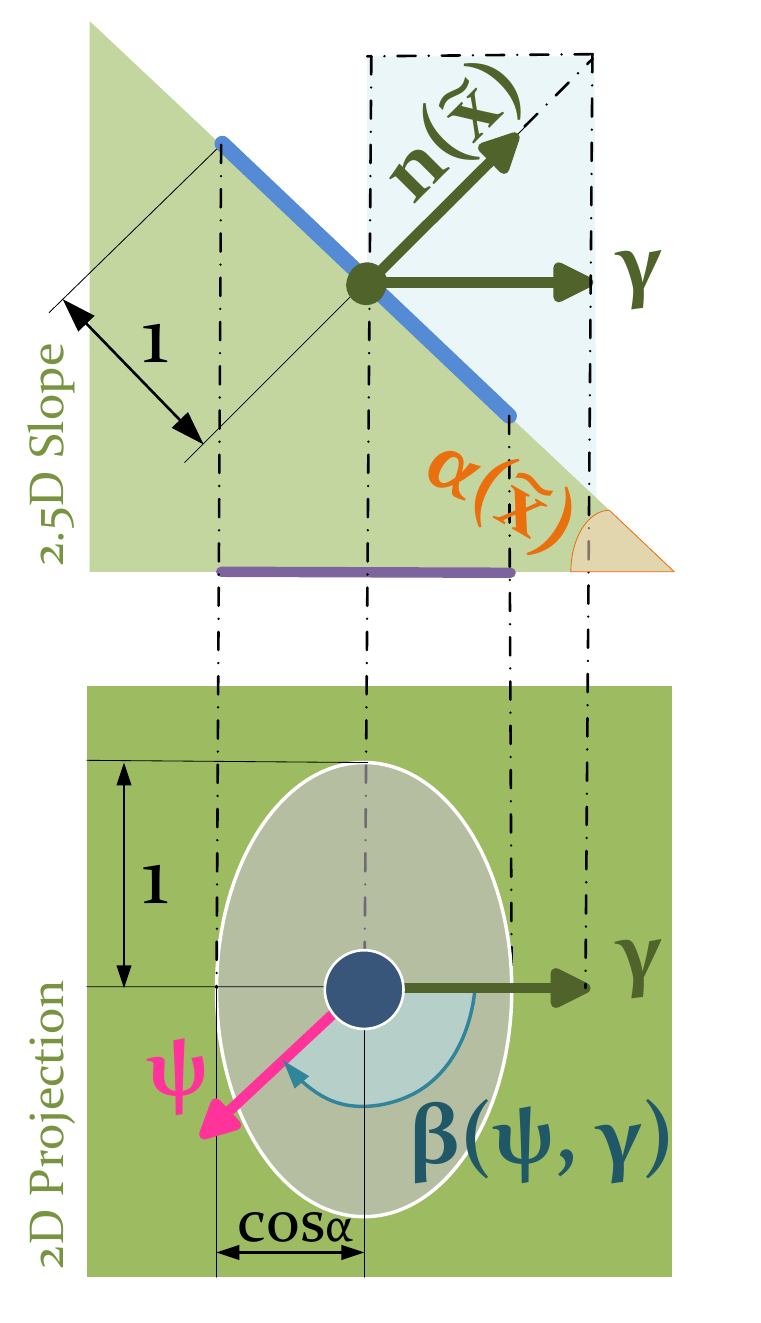}}
	\caption{Conceptual depiction of the slope model used to construct \textit{CAMIS}. Steepness $\alpha$, Aspect direction $\gamma$, Steering direction $u$ and Relative angle $\beta$ are indicated on it.}
    \label{fig:slopeParameters}
\end{figure}

The cost function $Q$, as seen in (\ref{eq:pathWaypoints}) and (\ref{eq:HJB}), determines the optimal path that results from \ac{bi-OUM}. In this section, we define the \textit{Continuous Anisotropic Model for Inclined Surfaces} or CAMIS cost function $C$, equivalent to $Q$ as shown in (\ref{eq:camisDefinition}).
\begin{equation}
\label{eq:camisDefinition}
    Q(\tilde{x},\psi) = C(\alpha(\tilde{x}),\beta(\psi(\tilde{x}),\gamma(\tilde{x})))
\end{equation}
Furthermore, this function models the behaviour of a mobile robot on irregular and unstructured terrain. It considers the gravity forces on the energetic performance of the robot, thus taking into account the existing slopes, defined by $\alpha$ and $\beta$ (as the abbreviation of $\alpha(\tilde{x})$ and $\beta(\psi(\tilde{x}),\gamma(\tilde{x}))$ respectively), within the scenario in question. Besides, it acknowledges as well the risks associated to high \textit{Pitch} and/or \textit{Roll} orientation angles due to traversing slopes.

\subsection{Modeling Slope-Robot Interaction}

The pose of a wheeled robot heavily depends on the shape of the terrain. The arrangement of all contact points between the wheels and the surface generally influences the orientation of the body frame. For this particular case, we introduce two simplifications. On the one hand, we consider a single contact point equivalent to all the previous ones, which serves as the origin of the body frame. On the other hand, we model the terrain for any location $\tilde{x}$ as a plane with an inclination determined by a parameter called \textit{Steepness} $\alpha(\tilde{x})$, as depicted in figure \ref{fig:slopeParameters}. 
The vector representing the direction of the slope, the aspect $\gamma(\tilde{x})$ (abbreviated as $\gamma$), is obtained by normalizing the projection of its normal vector $n(\tilde{x})$. Having this in mind, we
define the angle $\beta : \psi \times \gamma \to [-\pi, \pi] $ as the counter-clockwise rotation from $\gamma$ to $\psi$
, as seen in (\ref{eq:betavector}). 
\begin{equation}
\label{eq:betavector}
\beta(\psi,\gamma) = atan2\left([0,0,1] \cdot (\gamma \wedge \psi), \gamma \cdot \psi \right)
\end{equation}
This angle parameter
consequently takes a zero value when $\psi$ is coincident with $\gamma$. We name this particular orientation as the \textit{Descent} direction, which can be understood as well as the direction where the maximum fall step occurs. Its reverse direction takes the name of \textit{Ascent} direction, and therefore $\beta = \pm \pi$ for this case. Perpendicular to the aforementioned directions we find another two: the \textit{Lateral 1} ($\beta = \pi/2$) and \textit{Lateral 2} ($\beta = -\pi/2$) directions.

\subsection{Anisotropic Cost Function}
\begin{figure}[t]
    \centering
    \includegraphics[width=\columnwidth]{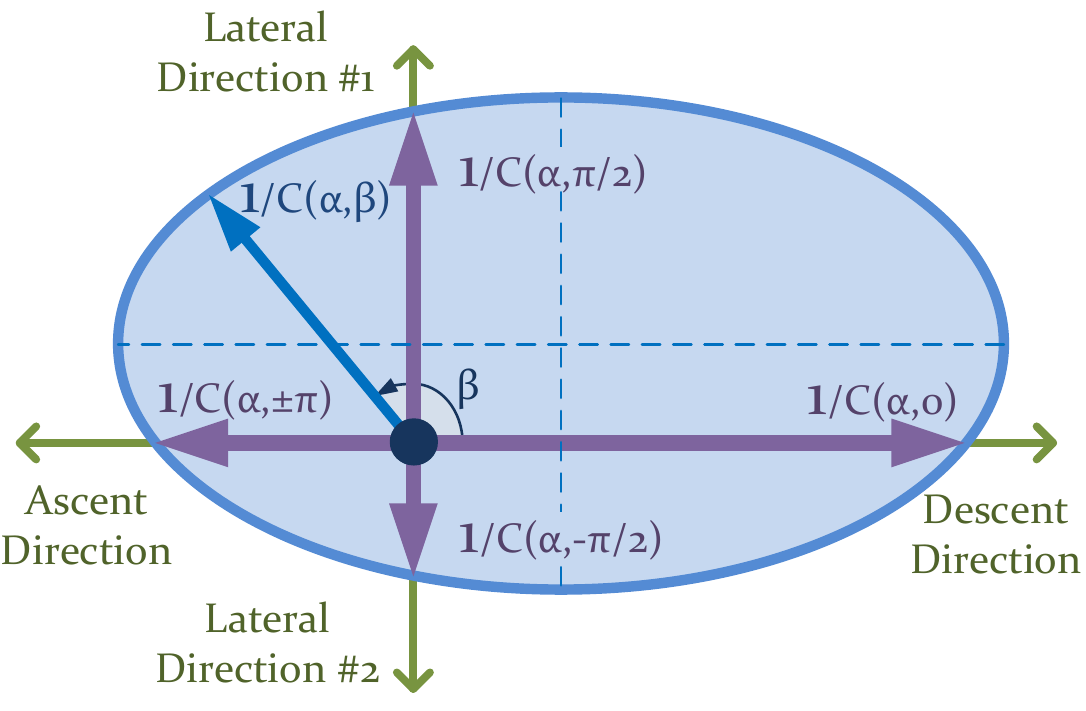}
    \caption{Geometric representation of the inverse cost function, corresponding to a displaced ellipse.}
    \label{fig:ellipsePropagationFunction}
\end{figure}

Thereafter, it is clear both $\alpha(\tilde{x})$ and $\beta(\psi(\tilde{x}),\gamma(\tilde{x}))$ not only model the slope placed at any location $x$ but also serve as inputs to cost function $C(\alpha, \beta)$. However, it still remains to be explained how the latter function is usable by anisotropic \ac{PDE} planners while also provides different values according to the angle $\beta$. The main idea behind this is that we force its inverse, which is analogous to the one in (\ref{eq:propagation}), to take the form of a closed conic curve. This geometrical shape is specifically a displaced ellipse as portrayed in figure \ref{fig:ellipsePropagationFunction}. By doing this, we make $1/C(\alpha,\beta)$ as result a closed, continuous, fully differentiable and convex function, and hence compatible with anisotropic \ac{PDE} planners requirements.
Then, we take the general polar form of a displaced ellipse to define $1/C(\alpha,\beta)$ as shown in (\ref{eq:ellipseGeneralForm}).
\setlength{\arraycolsep}{0.0em}
\begin{eqnarray}
\label{eq:ellipseGeneralForm}
 0 &{}={}& [\cos_\beta \sin_\beta]^T \left[ \begin{array}{cc} p_1(\alpha) & p_2(\alpha)/2 \\ p_2(\alpha)/2 & p3(\alpha) \end{array} \right] \left[ \begin{array}{c} \cos_\beta \\ \sin_\beta \end{array} \right] 1/C(\alpha,\beta)^2 + \nonumber\\ &&  
 + \left[ \begin{array}{cc} p_4(\alpha) & p_5(\alpha) \end{array} \right] \left[ \begin{array}{c} \cos_\beta \\ \sin_\beta \end{array} \right] 1/C(\alpha,\beta) + p_6(\alpha)
\end{eqnarray}
As can be noted, $1/C(\alpha,\beta)$ is constructed upon six other functions that, in turn, are defined over $\alpha$. In this way, the \textit{Steepness} $\alpha$ takes influence on the shape of the ellipse. A total of six functions
are necessary to fully define the displaced ellipse: $p_1(\alpha), p_2(\alpha), p_3(\alpha), p_4(\alpha), p_5(\alpha)$ and $p_6(\alpha)$.
By varying them based on $\alpha$, the length of the two ellipse axii, $a$ and $b$, as well as the displacement of the reference frame from its central point can change in consequence. Therefore, for a slope with certain values of $\alpha$ and $\gamma$ the function $1/C(\alpha,\beta)$ takes different values in function of $\beta$ (which, as seen in (\ref{eq:betavector}), results from the robot steering function $\psi$).
However, the use of these six functions that define the ellipse is counter-intuitive and their number may be reduced. Therefore, we opt to transform them into four cost functions where the value of $\beta$ is fixed to correspond to the four slope directions: \textit{Descent} ($\beta = 0$), \textit{Ascent} ($\beta = \pm \pi$), \textit{Lateral 1} ($\beta = \pi/2$) and \textit{Lateral 2} ($\beta = -\pi/2$).
By substituting them into (\ref{eq:ellipseGeneralForm}) we have the system and its solution in (\ref{eq:psystem}).
Moreover, in order to maintain the ellipse axes parallel to the Ascent-Descent and Lateral axes (as in figure \ref{fig:ellipsePropagationFunction}), we set $p_2(\alpha) = 0$. In this way, the ellipse is fully defined for each node of the grid by using just four cost values, which depend to its associated value of steepness $\alpha$.
\begin{equation}
\label{eq:psystem}
    \left[ \begin{array}{c} p_1(\alpha) / p_6(\alpha)  \\ p_2(\alpha)  / p_6(\alpha) \\ p_3(\alpha)  / p_6(\alpha)  \\ p_4(\alpha)  / p_6(\alpha)  \\ p_5(\alpha)  / p_6(\alpha)  \end{array}\right] = \left[ \begin{array}{c} - C(\alpha, \pm \pi) C(\alpha, 0) \\ 0 \\ - C(\alpha, -\pi/2) C(\alpha, \pi/2) \\ C(\alpha, \pm \pi) - C(\alpha, 0)  \\ C(\alpha, -\pi/2) - C(\alpha, \pi/2) \end{array}\right] 
\end{equation}

Finally, only remains for the path planner to know the value of the anisotropy coefficient according to the \textit{Steepness} $\alpha$ using expression (\ref{eq:anisotropyCoeff}) based on (\ref{eq:anisotropydefinition}) and (\ref{eq:camisDefinition}).
\begin{equation}
\label{eq:anisotropyCoeff}
\Upsilon(\alpha) = \frac{\max\limits_{\beta \in [-\pi, \pi]}  C(\alpha,\beta) }{\min\limits_{\beta \in [-\pi, \pi]}  C(\alpha,\beta) }
\end{equation}

\subsection{CAMIS complying with optimization criteria}

The next step is to relate $C(\alpha,\beta)$ to different factors associated with distinct physical phenomenons regarding the traverse of slopes. First, we start by introducing the \textit{Drawbar Pull Resistance Force} $R$, which models the power consumption demanded to drive a certain terrain according to the
\textit{specific resistance} coefficient $\rho$, \textit{mass} $m$ and \textit{gravity} $g$. An assumption we make here is that the robot speed $v$ is considered to be constant, and hence any inertia effects are ignored. First we start from
the model proposed by Rowe \textit{et al.}\cite{rowe1990optimal} to build $R = R(\alpha,\theta)$, where the angles of \textit{Pitch} $\theta$ and \textit{Steepness} $\alpha$ determine the energetic consumption of the robot, as in (\ref{eq:rowemodel}). This equation is the sum of the \textit{Rolling Resistance} \cite{perez2019choosing} and the pull produced by the gravity, divided by a factor of $\cos_{\theta}$ to account for the fact that we are solving a two dimensional path planning problem. In other words, the 2.5D elevation map is projected onto the 2D plane and the aforementioned robot speed $v$ takes different values in the 2D projection when climbing or descending through a slope, i.e. changing its Z-coordinate. This can be understood better by checking on figure \ref{fig:slopeParameters}, where the blue circle on the slope takes the form of an ellipse in its projection onto the XY-plane. 
\begin{equation}
\label{eq:rowemodel}
R(\alpha,\theta) = m g \left( \rho \frac{\cos_{\alpha}}{\cos_{\theta}} - \tan_{\theta} \right)
\end{equation}
\begin{equation}
\label{eq:pitchequation}
    \tan_{\theta} = \cos_{\beta} \tan_{\alpha}
\end{equation}
\begin{equation}
\label{eq:pitchequation2}
    \cos_{\theta} = \frac{\cos_{\alpha}}{\sqrt{\cos^2_{\beta} + \cos^2_{\alpha} \sin^2_{\beta}}}
\end{equation}

Thereafter, we combine equations (\ref{eq:pitchequation}) and (\ref{eq:pitchequation2}), which geometrically relate \textit{Pitch} angle $\theta$ with $\alpha$ and $\beta$, with (\ref{eq:rowemodel}). The resulting expression is the baseline for $C(\alpha,\beta)$, which is expressed in equation (\ref{eq:energyriskC}).
Here, $C(\alpha,\beta)$ accounts for the effect of braking while descending through a slope, and incorporates the \textit{Bezier} function $R_b$, the slip ratio $s_r = s_r(\alpha)$, the slip angle $s_a = s_a(\alpha)$ and the \textit{Roll} weight function $w_{\phi} = w_{\phi}(\alpha)$. Each of these modifications are justified next.
\begin{equation}
\label{eq:energyriskC}
    C(\alpha, \beta) = \begin{cases}
    \Biggl| \sqrt{\left( \frac{(\rho + \tan_{\alpha} + R_b)\cos_{\beta}}{2(1 - s_{r})}\right)^2 +  \left(\frac{\rho \cos_{\alpha} w_{\phi} \sin_{\beta}}{\cos_{s_{a}}}\right)^2} - \\ \hfill \hfill  \frac{\rho + \tan_{\alpha} - R_b} {2} \cos_{\beta} \Biggr|\frac{m g}{v} \ , \\ \hfill \hfill \arctan_{\rho} - \alpha_{\Delta} < \alpha < \arctan_{\rho} + \alpha_{\Delta}
      \\ \\  \Biggl| \sqrt{\left(\frac{\rho^2  \cos_{\beta}}{1 - s_{r}}\right)^2 +  \left(\frac{\rho \cos_{\alpha} w_{\phi} \sin_{\beta}}{\cos_{s_{a}}}\right)^2} - \\ \hfill \hfill  \tan_{\alpha} \cos_{\beta} \Biggr|  \frac{m g}{v} \ , \\ \hfill \hfill otherwise
    \end{cases}
\end{equation}

\begin{figure}
    \centering
    \includegraphics[width=\columnwidth,align=c]{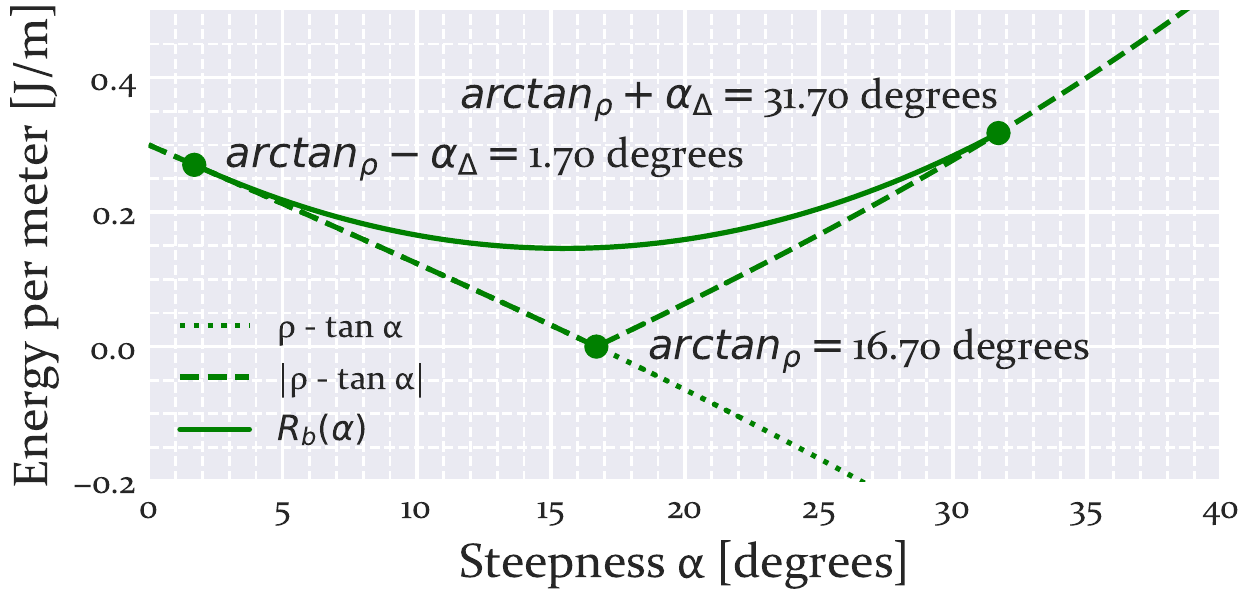}
    \caption{Use of Bezier curve $R_b$ to comply with the non-zero positive cost specification in a smooth way. $\rho = 0.3$ and $\alpha_{\Delta} = 15deg$ are used.}
    \label{fig:brakingEffect}
\end{figure}

With regards the braking effect, when $\alpha \geq \arctan_{\rho}$, zero or even negative values can appear in the energy consumption as shown in figure \ref{fig:brakingEffect}. This is mainly because in this case, the vehicle starts accelerating if no effort to counteract it is performed. To avoid this, the robot can start braking to keep the same speed, which we consider by applying the absolute operation on (\ref{eq:energyriskC}). However, to comply with the PDE planner requirements, it is still needed to overcome the problems of having zero values and a discontinuity. We propose hence the use of a quadratic \textit{Bezier} curve $R_b(\alpha)$ to overcome this problem, having the point where $\alpha = \arctan_{\rho}$ as one control point and the other two at $\alpha = max(0,\arctan_{\rho} - \alpha_{\Delta})$ and $\alpha = \arctan_{\rho} + \alpha_{\Delta}$, being $\alpha_{\Delta}$ a certain margin value. In this way not only zero cost values are avoided, but also a certain margin is left and discontinuities are prevented.

Regarding the slippage, it is expressed  using two variables: the slip ratio $s_{r} = s_{r}(\alpha)$ and the slip angle $s_{a} = s_{a}(\alpha)$ \cite{ishigami2011path}. The first comes as a function of the pitch angle, while the second does the same with the roll angle. Therefore, the slip ratio will influence more in the ascent-descent direction, while the slip angle will do this in the lateral ones. With regards $w_{\phi}$, as is demonstrated later in the tests it allows to control the angle of attack to traverse slopes, i.e. to penalize or not \textit{Roll} angle $\phi$ as consequence. This may prove useful to penalize \textit{Roll} angles that can jeopardize the stability of the robot and to prevent lateral turn-over. Finally, the constant robot speed $v$ is introduced as well into the cost function in equation (\ref{eq:energyriskC}), leaving it expressed as result in terms of energy consumption per meter.

\section{Results}
\label{sec-testing}
This section presents the outcome of the use of CAMIS in path planning.
We prepared and carried out two numerical simulations and a field test. 
As regards the simulation tests, the Python scripts used to run them as well as to implement \textit{bi-OUM} and \textit{CAMIS} are published online to a public GitHub repository\footnote{\url{https://github.com/spaceuma/CAMIS_python}}. The main idea behind these tests is to exhaustively analyze how different configurations of \textit{CAMIS}, either prioritizing the minimization of energy or to prevent the \textit{Roll} angle from taking high values, determine the shape of the resulting paths. Besides, the benefits and drawbacks of using direction-dependent (anisotropic) methods are disclosed in the simulations and compared to isotropic methods, being the latter compatible with \textit{Eikonal}-exclusive solvers such as the computationally cheaper Fast Marching Method (FMM).

In the case of the field test, we demonstrate how \textit{CAMIS} can be taken to reality. We introduce the information regarding the setup used, consisting of an experimental skid-steering robot and a real terrain that contains slopes. Finally, we present a comparative between the performance expected by the plan and the real robot in the field.

\subsection{Energy Optimization}
\begin{figure}
    \centering
    \subfloat[Slip Ratio.
        \label{fig-slipRatio}]%
    	{\includegraphics[width=0.48\columnwidth,align=c]{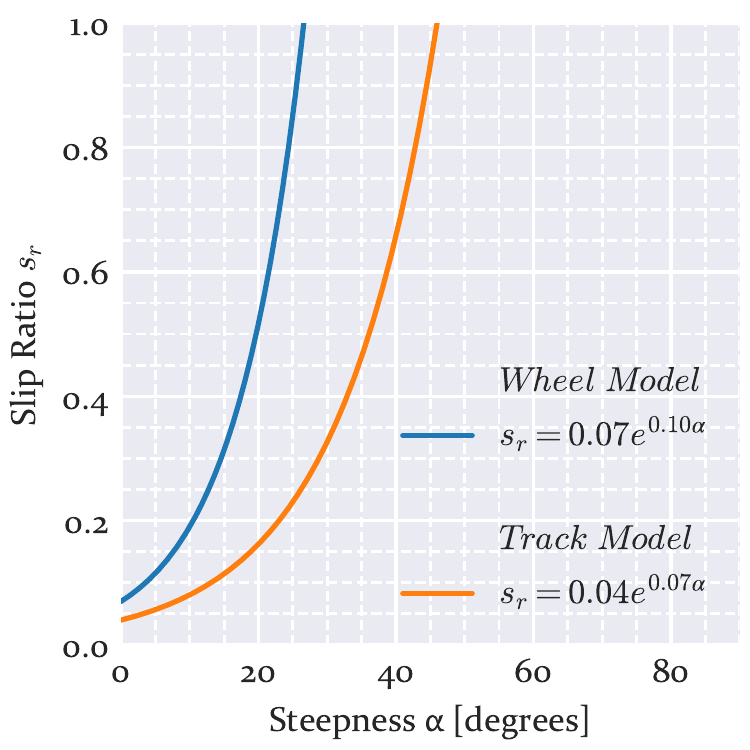}}
    \quad
	\subfloat[Slip Angle.
	\label{fig-slipAngle}]%
    	{\includegraphics[width=0.48\columnwidth,align=c]{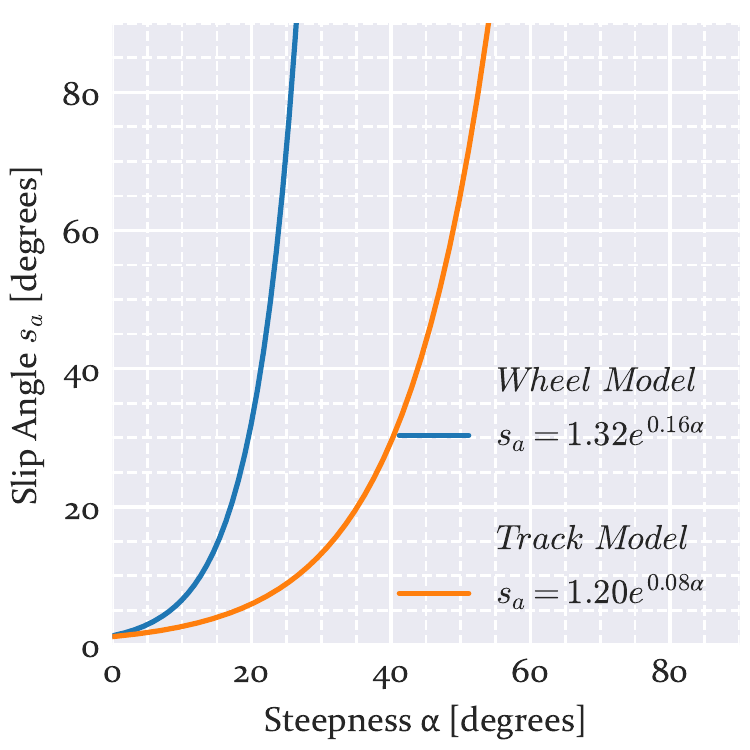}}
	\caption{Slip angle and slip ratio functions of the two models (\textit{Wheel} and \textit{Track}) found in the literature, used in the numerical simulation tests.}
    \label{fig-slipFunctions}
\end{figure}

The first test serves to investigate the influence of the functions and parameters that make up \textit{CAMIS} and affect the robot energy consumption: the parameter \textit{Specific Resistance} $\rho$ and the functions slip ratio $s_r$ and the slip angle $s_a$.
With regards the mass $m$ and the gravity $g$, they are directly proportional to the cost function in all heading directions, and hence they have no impact on the shape of the resulting paths. This means they will only proportionally increase or decrease the value of \textit{Total Cost} that entails any optimal path, but without changing the path itself. The \textit{Roll} weight function $w_{\phi}$ is omitted in this case by making it return a constant value of one.
\begin{figure*}
    \centering
    \includegraphics[width=1.0\textwidth,align=c]{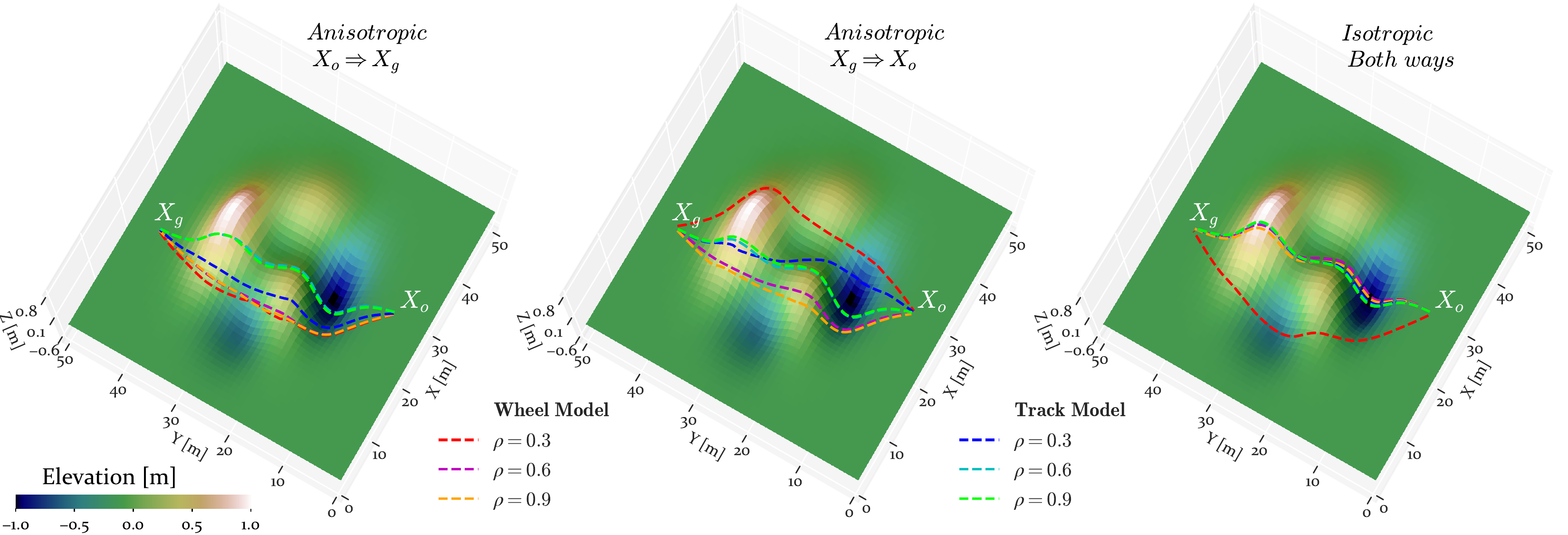}
    \caption{3d view of the paths resulting from the first simulation test.}
    \label{fig:numericalPaths}
\end{figure*}

On the one hand, the \textit{specific resistance} coefficient $\rho$ admits values between 0 and 1. For this test we opted for choosing 0.3, 0.6 and 0.9. On the other hand, to define the slip functions in a representative and meaningful way we make use of two models found in the bibliography \cite{sutoh2015right}. Both of these models were created after the results of some experiments using two locomotion mechanisms, a wheel and a track. Both mechanisms were put into action to drive on top of an inclined sandbox filled with lunar regolith simulant \cite{wakabayashi2009design}. The resulting slip ratio $s_r$ and slip angle $s_a$ functions fitting the obtained data are depicted in figures \ref{fig-slipRatio} and \ref{fig-slipAngle} respectively. It can be observed how the \textit{Wheel} model returns higher values than the \textit{Track} model in both functions as the \textit{Steepness} $\alpha$ increases. Besides, there are values of $\alpha$ in which these functions produce discontinuities in the form of divisions by zero in (\ref{eq:energyriskC}). This is the case when slip ratio $s_r$ takes a value of 1 and the slip angle reaches a value of 90 degrees. This issue proved to be very problematic in preliminary tests since the continuity in the \textit{CAMIS} function was broken as a consequence. Although a possible workaround would be to assign a relatively high value of cost whenever this happens, the anisotropy produced would be still too high (around 10.0 or more), specially due to the fact that the values of $\alpha$ producing a discontinuity can be different for $s_r$ and $s_a$. As a result, unless a more refined grid is used, which entails an increase in the number of nodes, the path planner could produce local minimal points in the computed values of \textit{Total Cost} $T$, which in turn is fatal for later extracting the path. Having this in mind, we make the simulation tests in a scenario where the slip functions do not produce discontinuities, but we remark this issue and leave it to be investigated in future work.
\begin{figure}
    \centering
    \subfloat[Values of \textit{Steepness} $\alpha$.
	\label{fig-numericalSlope}]%
    	{\includegraphics[width=0.48\columnwidth,align=c]{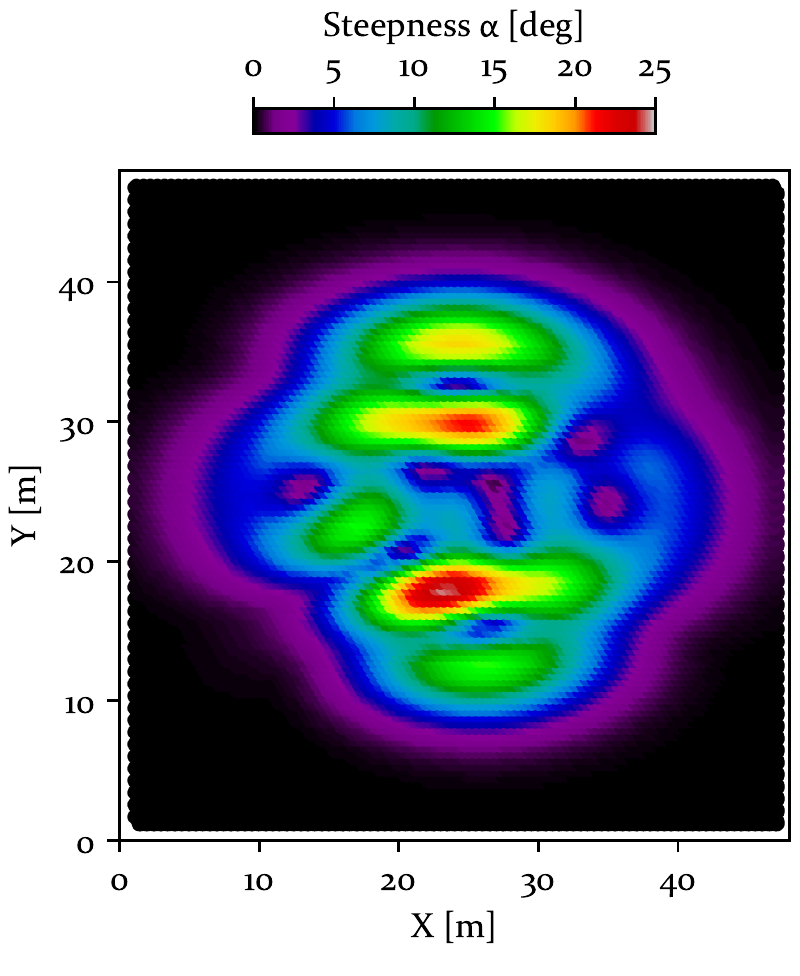}}
    \quad
	\subfloat[Values of aspect angle.
        \label{fig-numericalAspect}]%
    	{\includegraphics[width=0.48\columnwidth,align=c]{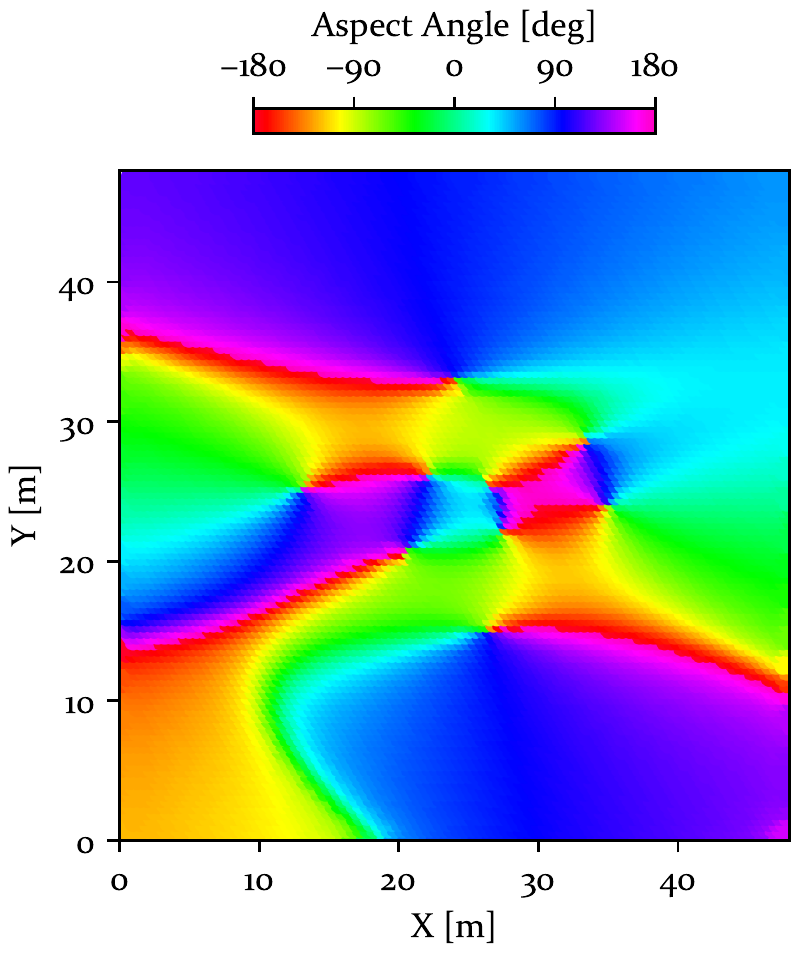}}
	\caption{Slope data describing the shape of the terrain used in the first simulation tests.}
    \label{fig-numericalMaps}
\end{figure}
\begin{figure}
    \centering
    \includegraphics[width=1.0\columnwidth,align=c]{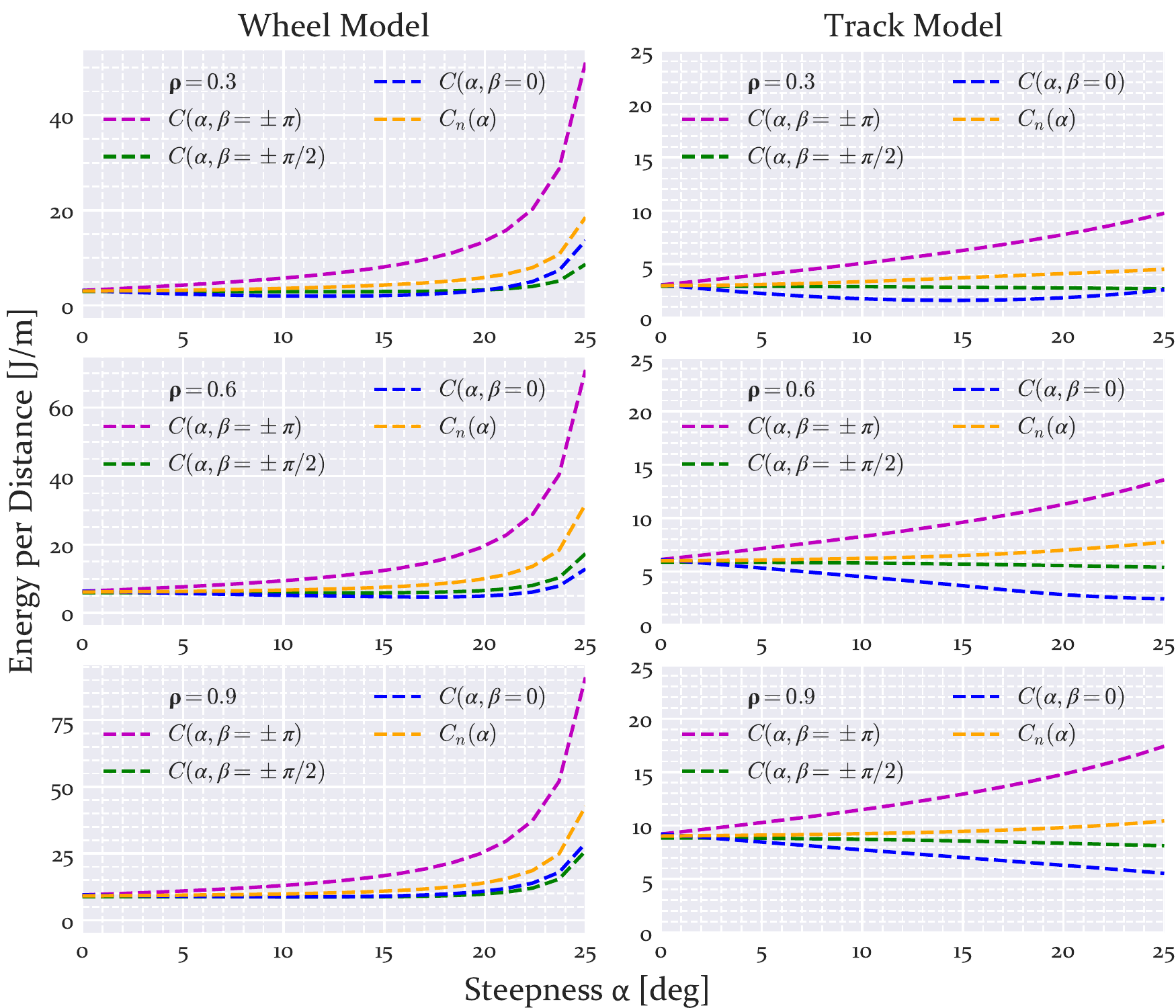}
    \caption{Values of cost returned by each anisotropic model (and the corresponding isotropic equivalent $C_n$) according to \textit{Steepness} $\alpha$.}
    \label{fig:numericalCosts}
\end{figure}

The terrain used in this test is created by hand with the idea of making it contain multiple kinds of slopes and see how they affect the path planner. Figure \ref{fig:numericalPaths} portrays, together with the resulting paths that will be explained later, a 3d representation of the custom DEM, a 50x50m square grid with 0.1m of resolution containing elevation values. This DEM is later converted into an hexagonal grid with 0.5m of resolution for the anisotropic path planner. As mentioned, it contains several hills and craters irregularly arranged in different locations and shapes. 
A more detailed description about the characteristics of the existing slopes is graphically provided in Figure \ref{fig-numericalMaps}. The first plot, figure \ref{fig-numericalSlope}, provides an indication of the value of \textit{Steepness} that is present at each location. Figure \ref{fig-numericalAspect} does the same but in this case with the \textit{Aspect} angle, which takes a value of zero degrees when the slope faces a direction parallel to the X-axis in increasing values. It can be noted how this terrain presents smooth height variations and a variety of \textit{Steepness} values ranging from 0 to near 25 degrees, which means in this situation the slip models used will not create the discontinuities mentioned before.

The paths produced and shown in figure \ref{fig:numericalPaths} connect two positions, $X_o$ and $X_g$, using different configuration sets that combine the aforementioned values of $\rho$ together with the two slip models: the \textit{Wheel} model and the \textit{Track} model. Figure \ref{fig:numericalCosts} depicts a series of plots showing the values returned by $C(\alpha,\beta)$ in each configuration, according to the \textit{Steepness} $\alpha$ and given certain values of $\beta$. As can be denoted, the cost increases rapidly in the case of the \textit{Wheel} model, since as we mentioned the slip ratio gets closer to one. This makes the paths produced by this model get further from slopes, surrounding the hills and sinking in the middle. Due to the differences in \textit{Steepness} $\alpha$ and aspect, seen in figures \ref{fig-numericalSlope} and \ref{fig-numericalAspect}, these paths take different shapes in the first go and the return, specially when $\rho = 0.3$. In the case of those paths produced by the \textit{Track} model, they pass on top of the slopes. This is because the differences in cost, not only according to direction but also according to \textit{Steepness} $\alpha$, are not that much significant. To make the comparative with paths produced by isotropic cost functions, we made functions of this kind equivalent to the anisotropic ones. We made this by setting a nominal cost $C_n(\alpha)$ that only depends on the \textit{Steepness}. The main idea is that $C_n(\alpha)$, instead of being the inverse of an ellipse, is the inverse of a circumference (which in turn produces another circumference). The radius of such circumference, i.e. the value returned by $C_n(\alpha)$, is chosen so the area enclosed is equal to the area of the ellipse in the anisotropic case. In the isotropic case, as seen in figure \ref{fig:numericalPaths}, the paths produced by $C_n(\alpha)$ are the same for going from $X_o$ to $X_g$ and going from $X_g$ and $X_o$. 
\begin{figure}
    \centering
    \includegraphics[width=1.0\columnwidth,align=c]{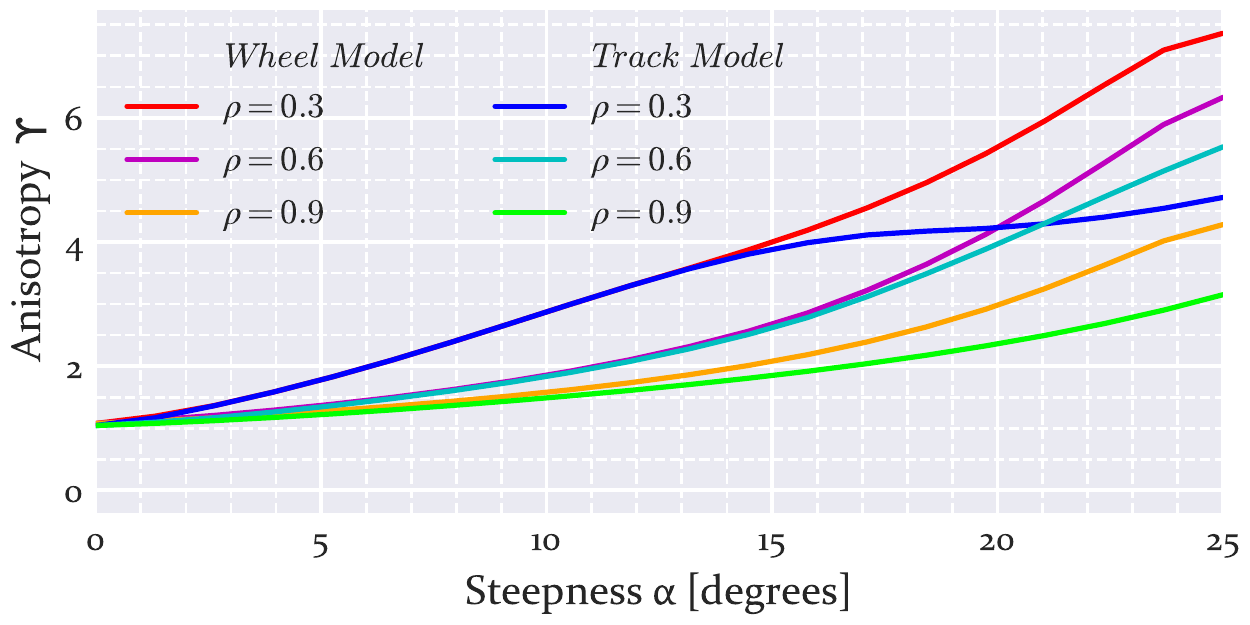}
    \caption{Values of anisotropy according to each slip model and the value of $\rho$ chosen.}
    \label{fig:numericalAnisotropies}
\end{figure}

\begin{figure}
    \centering
    \includegraphics[width=1.0\columnwidth,align=c]{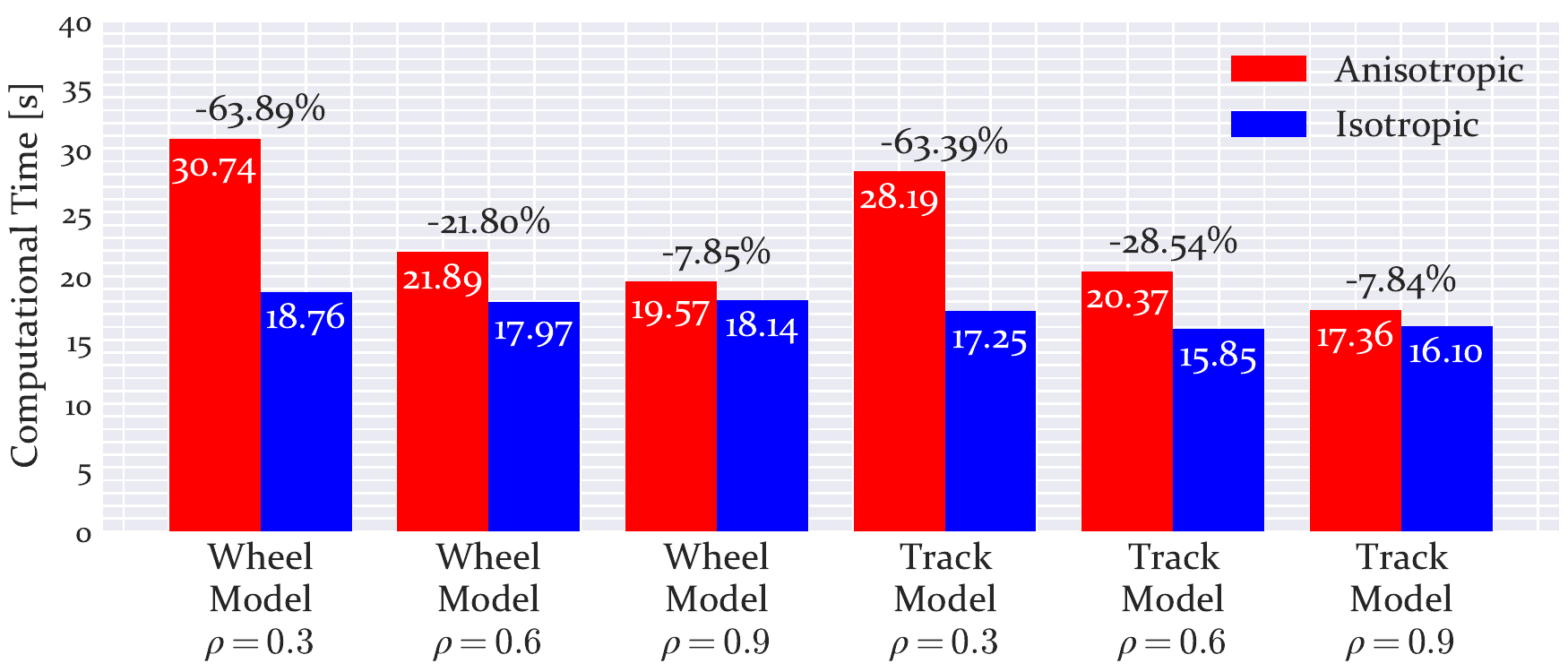}
    \caption{Comparative of computational times between anisotropic and isotropic functions.}
    \label{fig:numericalTimes}
\end{figure}

\begin{figure}
    \centering
    \includegraphics[width=1.0\columnwidth,align=c]{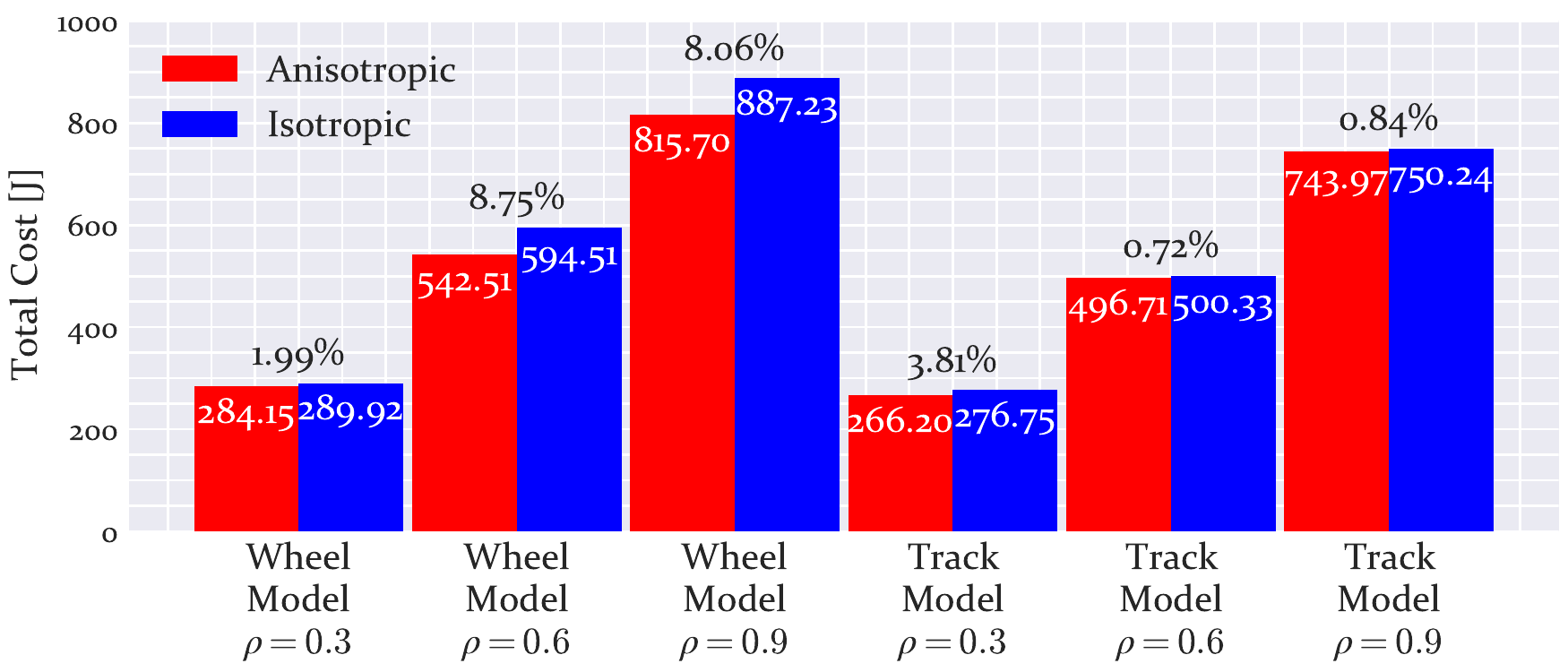}
    \caption{Estimated total cost (energy) for each configuration, together with the difference in percentage between the anisotropic and isotropic cases.}
    \label{fig:numericalResults}
\end{figure}

The anisotropy that results from the different configurations is displayed in figure \ref{fig:numericalAnisotropies}. In general all configurations tend to increase the anisotropy as the \textit{Steepness} $\alpha$ increases, but at different rates. This results on the map presenting more or less anisotropy depending on how the \textit{Steepness} is arranged, which in turn affects the time invested on computing the \textit{Total Cost}. This is mainly because the anisotropy of a certain node determines how many nearby nodes are considered to compute the corresponding value of \textit{Total Cost}, i.e. how many times an update of this value is made as previously denoted in figure \ref{eq:updateT}. This correlation is noticed in figure \ref{fig:numericalTimes}, where, having executed the test on an Intel Core i7 processor with 3.40 GHz, two cases spent the most time: the \textit{Wheel} Model and the \textit{Track} Model both at $\rho = 0.3$. 
As there are many nodes with a \textit{Steepness} between 5 and 15 degrees (see figure \ref{fig-numericalSlope}), and these two cases have high anisotropy between those values (see figure \ref{fig:numericalAnisotropies}), it is straightforward to understand why they spent much more time in the computation. In fact the case of the \textit{Track} Model takes less time since the anisotropy does not increase much in values of \textit{Steepness} higher than 15 degrees. However, high anisotropy does not imply that the reduction in \textit{Total Cost} with respect to the equivalent isotropic model will be relatively high as well, i.e. the difference may not be much notorious.
As can be checked in figure \ref{fig:numericalResults}, the percentage of \textit{Total Cost} with respect to the isotropic case is between almost 2 and 4 percent. In the other two cases, the \textit{Wheel} model for $\rho = 0.6$ and $\rho = 0.9$, the difference produced by opting for using \textit{CAMIS} reaches almost 9 percent, a quite significant value in exchange of the increase in computational time indicated in figure \ref{fig:numericalTimes}. Nevertheless, for the same values of $\rho$, this difference is almost negligible in the case of the \textit{Track} model. The main reason is because achieving a bigger percentage can only be obtained not just by having high anisotropy but also by a cost function that, in average, returns much higher values as the \textit{Steepness} does. With this in mind, we can conclude that the main actor that justifies the use of \textit{CAMIS} to purely minimize energy consumption instead of isotropic methods is the slippage (in the form of slip angle $s_a$ and slip ratio $s_r$), as the slip makes the \textit{Cost} substantially increase in all directions. For this reason, the \textit{Wheel} model, suffering more from slippage than the \textit{Track} model, has in most cases more difference in the \textit{Total Cost} between the anisotropic and isotropic models than the second.


\begin{figure}
    \centering
    \subfloat[Panoramic view of the slope considered for the tests.
        \label{fig-realCuesta01}]%
    	{\includegraphics[width=\columnwidth,align=c]{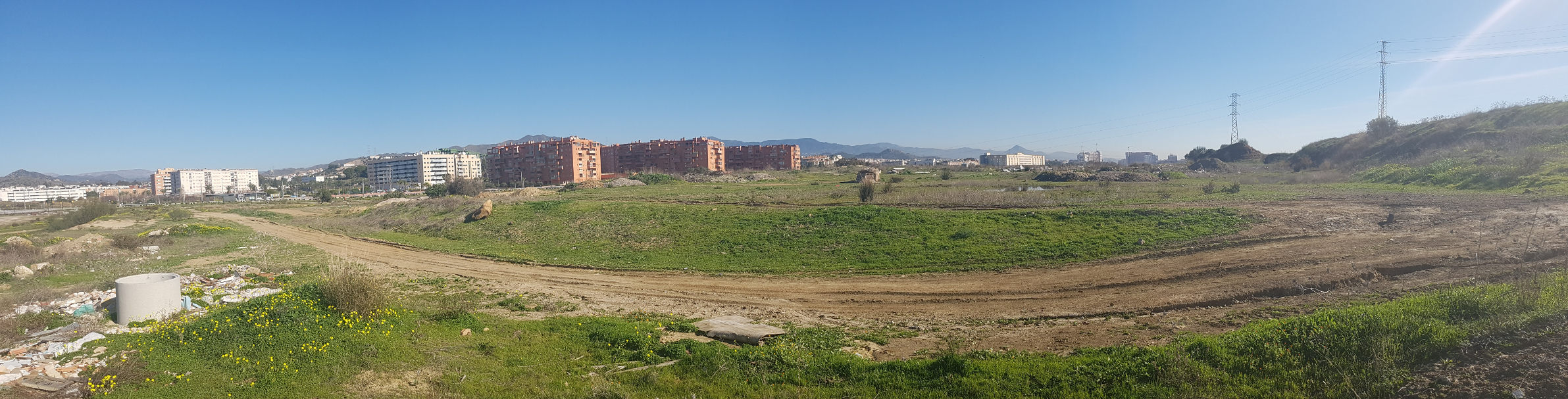}}
    \quad
    \subfloat[View of half of the slope.
        \label{fig-realCuesta02}]%
    	{\includegraphics[width=0.48\columnwidth,align=c]{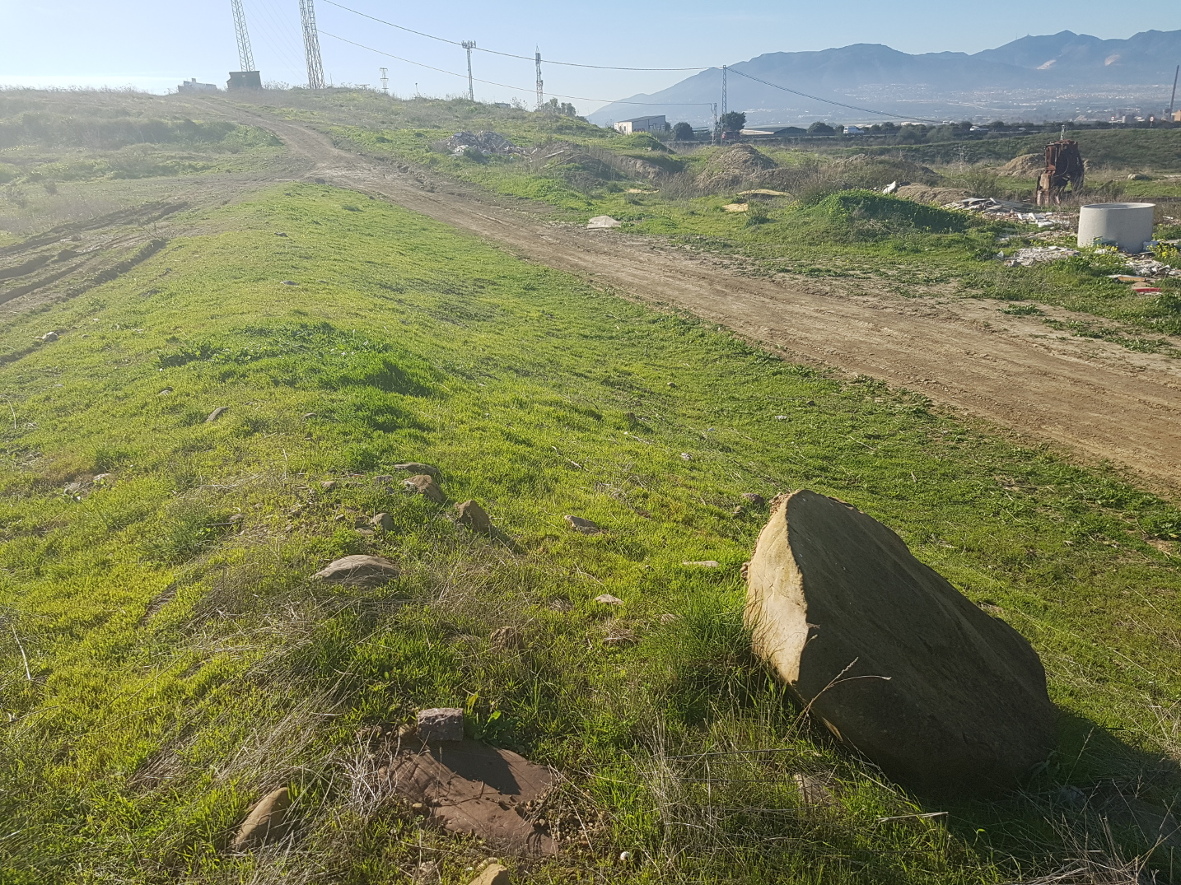}}
    \quad
    \subfloat[Another view from the other side.
        \label{fig-realCuesta03}]%
    	{\includegraphics[width=0.48\columnwidth,align=c]{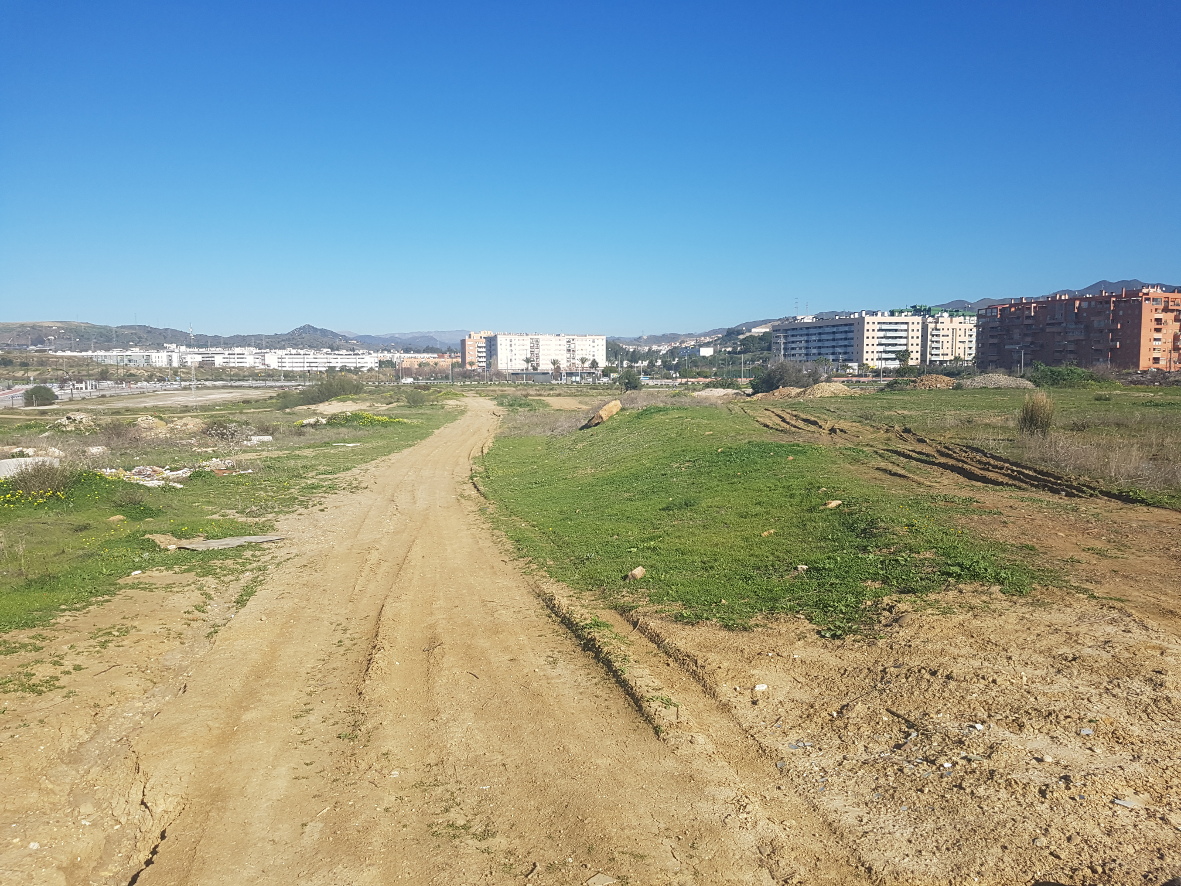}}
    \quad
    \subfloat[Screenshot showing the 3d model built using the Pix4Dmapper software, together with red arrows indicating the location of the four points of interest used in the tests.
        \label{fig-realCuesta3d}]%
    	{\includegraphics[width=\columnwidth,align=c]{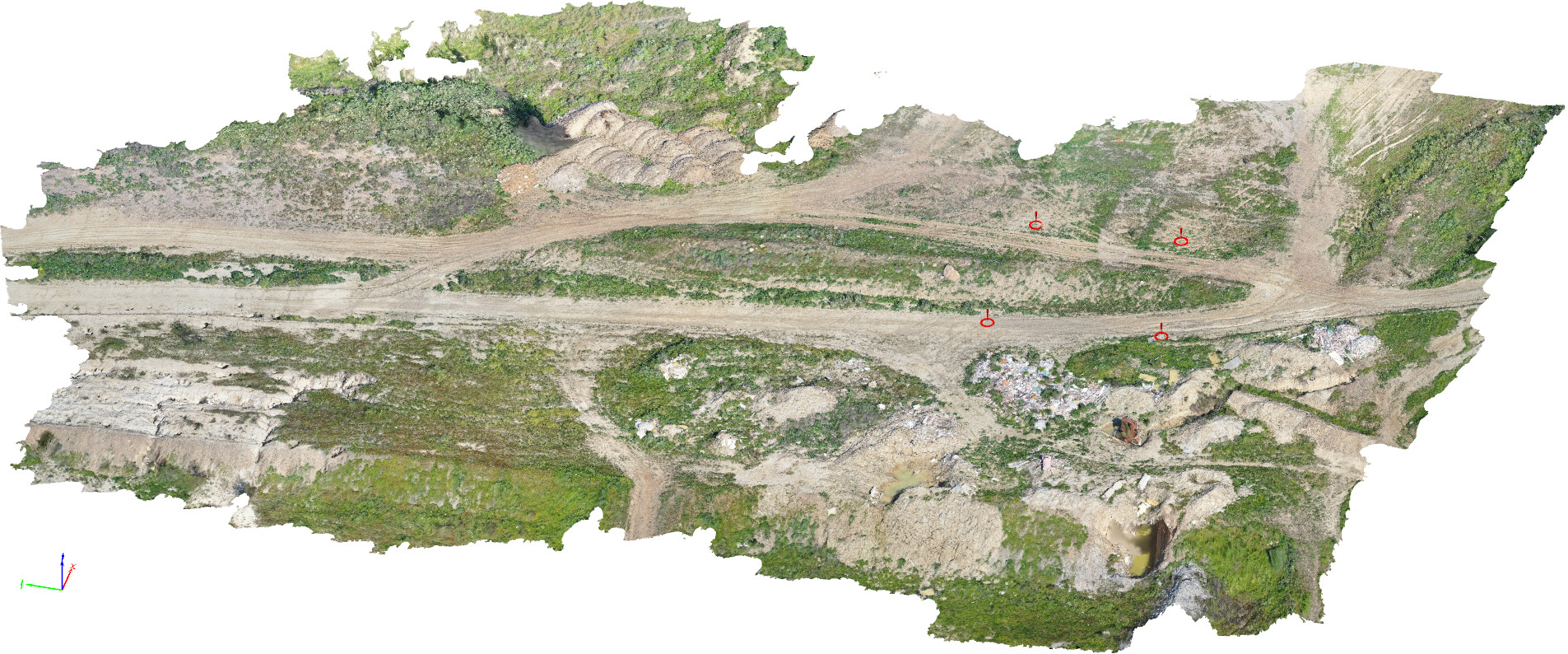}}
	\caption{Showcase of pictures showing the terrain using different perspectives (a)(b)(c) and a screenshot of Pix4Dmapper showing its virtual reconstruction.}
    \label{fig-umaCuesta}
\end{figure}
\begin{figure}
    \centering
    \subfloat[Frontal view of \textit{Cuadriga} robot.
        \label{fig-cuadrigafront}]%
    	{\includegraphics[width=0.48\columnwidth,align=c]{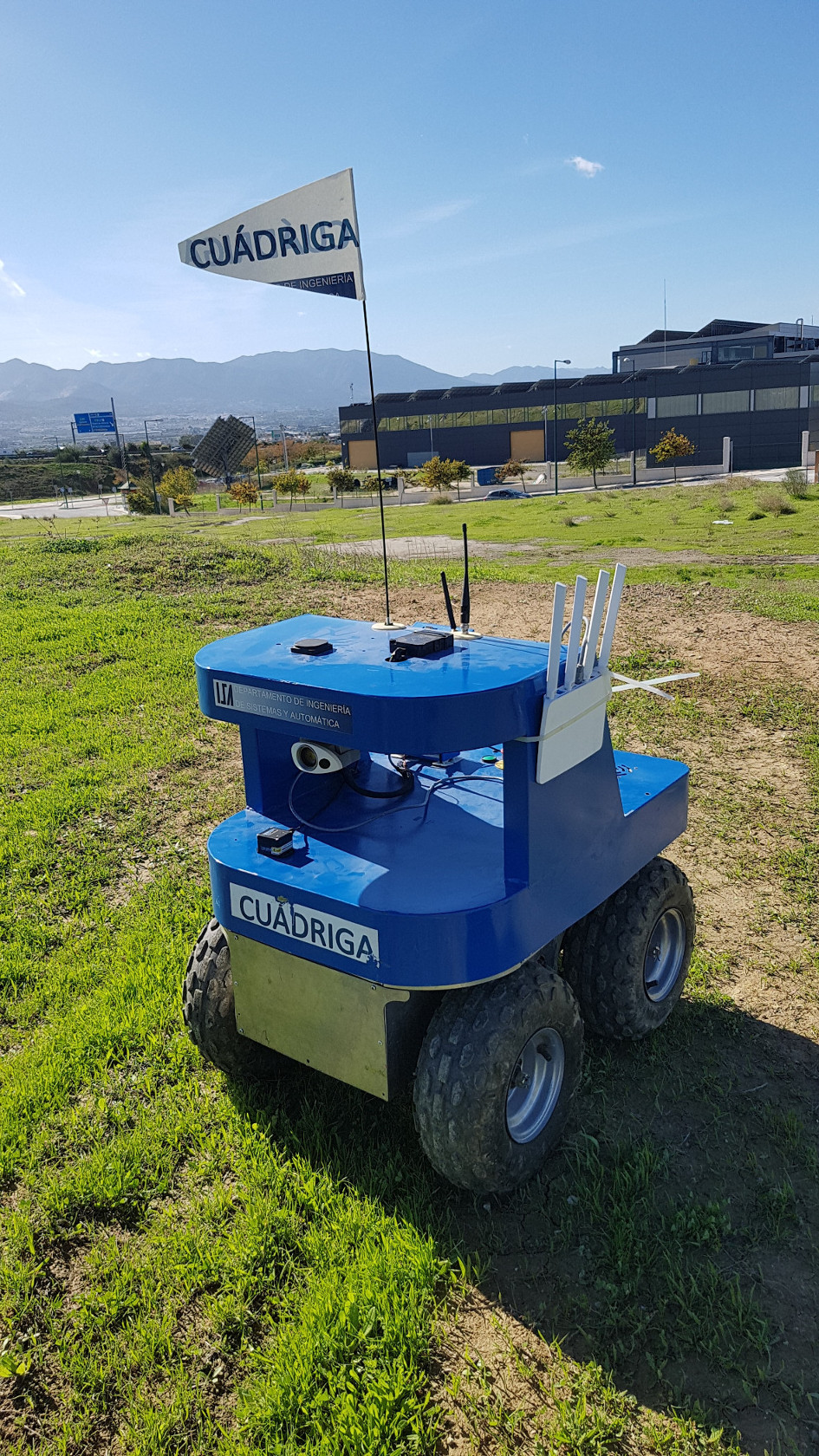}}
    \quad
    \subfloat[Rear view of \textit{Cuadriga} robot.
        \label{fig-cuadrigarear}]%
    	{\includegraphics[width=0.48\columnwidth,align=c]{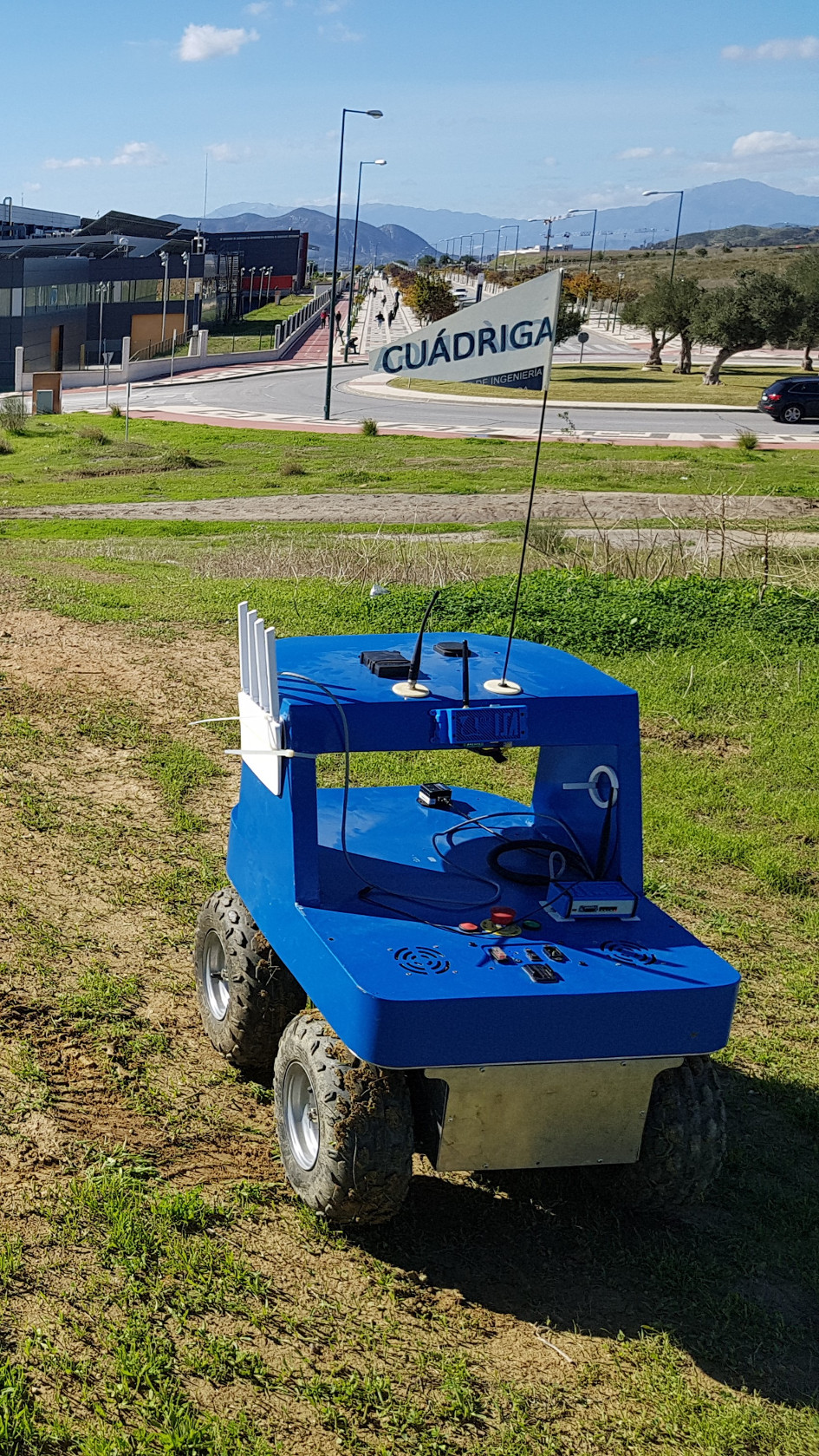}}
	\caption{The skid-steering four-wheeled \textit{Cuadriga} robot.}
    \label{fig-cuadrigaData}
\end{figure}
\begin{figure}
    \centering
    \subfloat[\textit{Cuadriga} climbing a slope different from the one used in the tests.
        \label{fig-cuadrigafront}]%
    	{\includegraphics[width=\columnwidth,align=c]{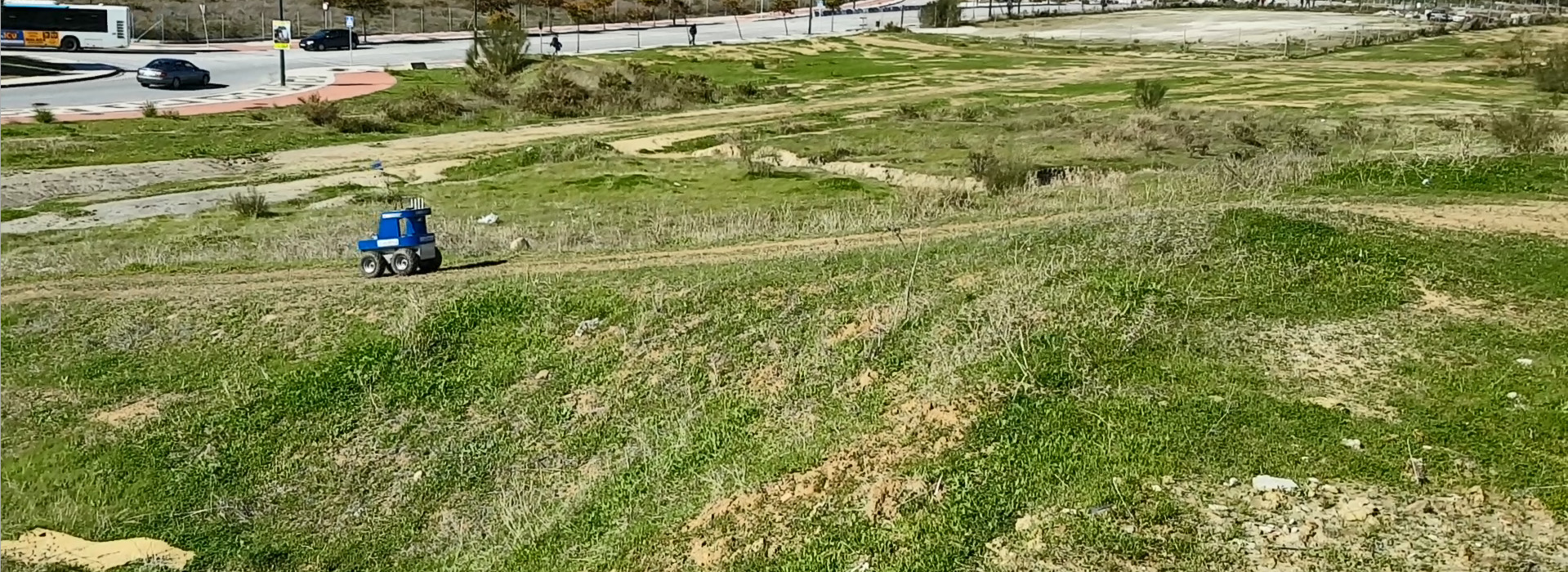}}
    \quad
    \subfloat[\textit{Cuadriga} descending the same slope.
        \label{fig-cuadrigarear}]%
    	{\includegraphics[width=\columnwidth,align=c]{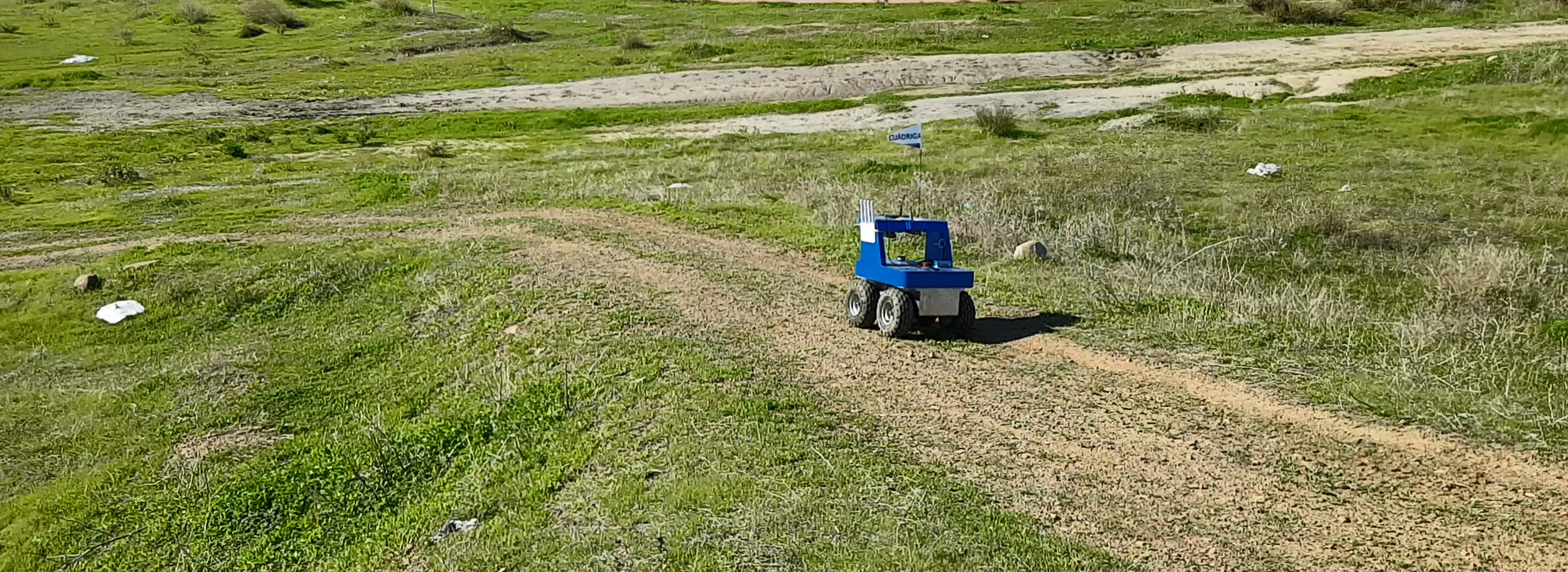}}
    \quad
	\subfloat[Energy per distance from preliminary drives.
	\label{fig-adCuadrigaCostFunction}]%
    	{\includegraphics[width=\columnwidth,align=c]{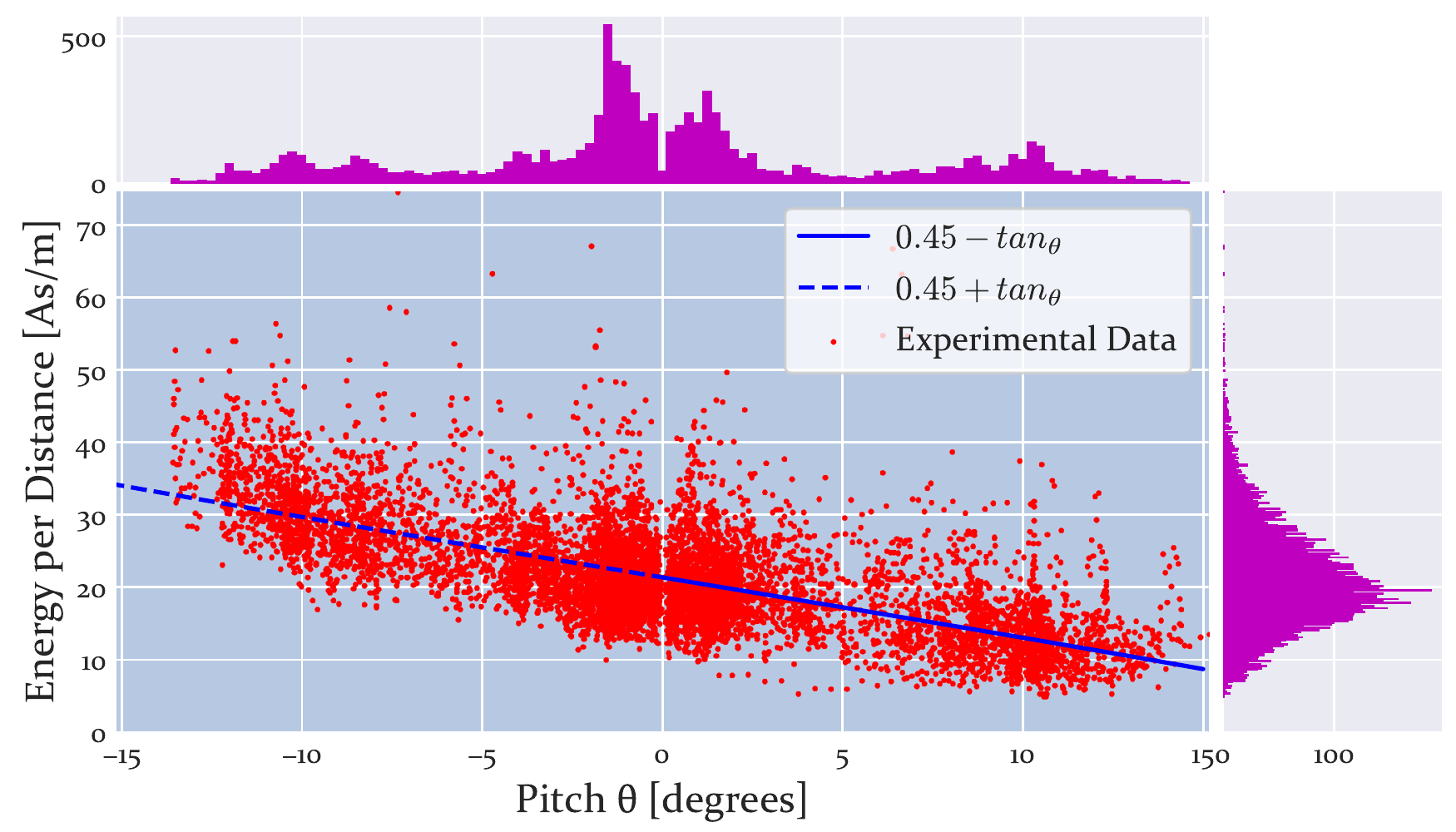}}
    \quad
    \subfloat[Slip Ratio from preliminary drives.
        \label{fig-adCuadrigaSlipFunction}]%
    	{\includegraphics[width=\columnwidth,align=c]{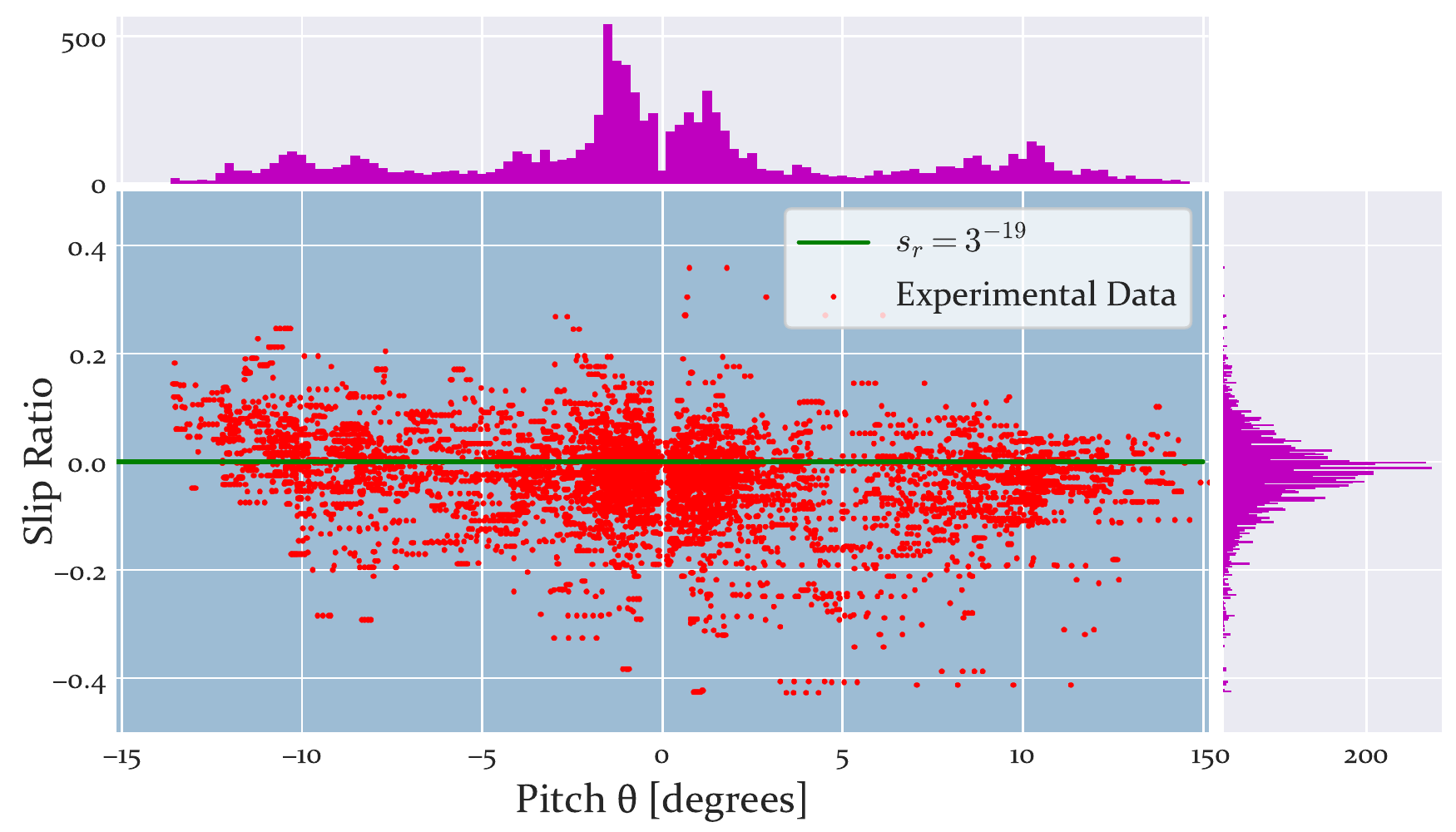}}
	\caption{Extraction of $\rho$ and $s(\alpha)$ based on preliminary drives of \textit{Cuadriga} going through different slopes that are made of the same ground material as the slope used in the tests.}
    \label{fig-cuadrigaData}
\end{figure}

\begin{figure*}
    \centering
    \includegraphics[width=1.0\textwidth,align=c]{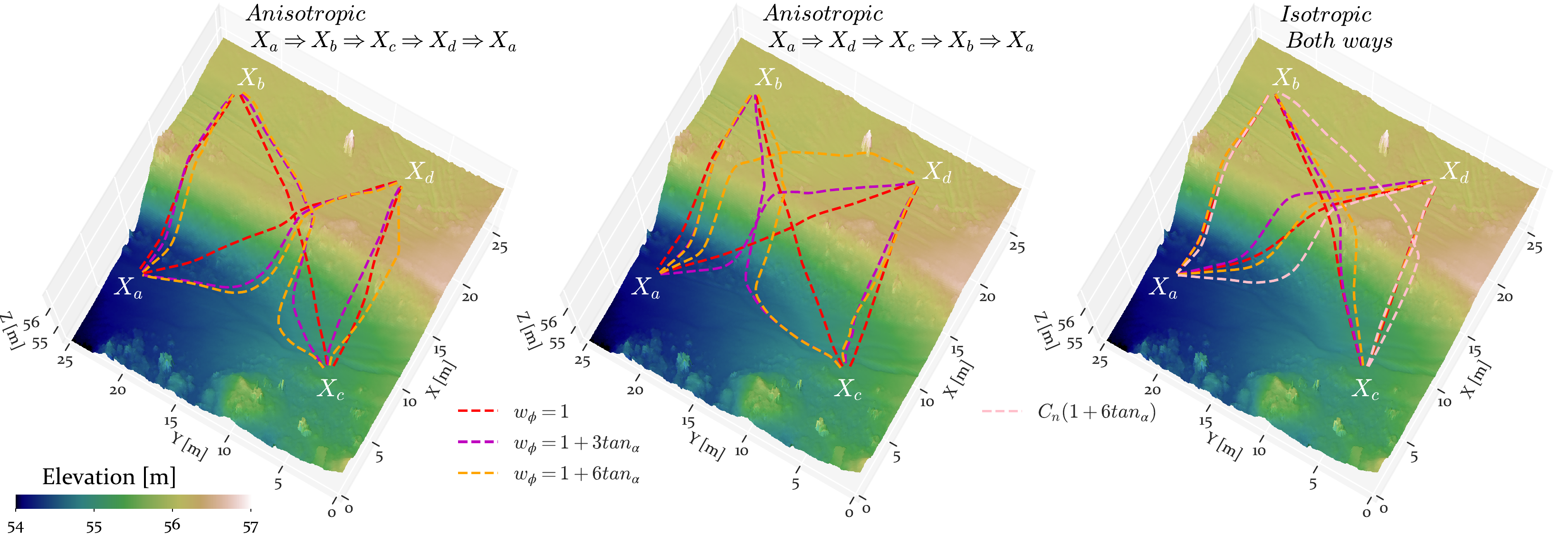}
    \caption{Paths traversing the slope with certain configurations of weight values.}
    \label{fig:roll_test_paths}
\end{figure*}

\subsection{Experimental Setup}
For the upcoming experiments, consisting of a second simulation and a field test, we made use of a DEM that describes the shape of an experimental terrain thanks to the use of photogrammetry software. Figure \ref{fig-umaCuesta} displays some pictures showing this terrain, which is a portion of a dedicated $90000m^2$ outdoor experimental area at the University of Malaga campus. This is an unstructured natural environment with different terrain altitudes that is used to perform large-scale disaster response exercises organized by the Chair of Safety, Emergencies and Disasters at the University of Malaga \cite{mandow2020experimental}. 
The chosen portion of terrain contains a slope in the middle that divides it into two levels of height. The 3d virtual model that is shown in figure \ref{fig-realCuesta3d} was constructed using the Pix4Dmapper software \cite{barbasiewicz2018analysis} with version 4.6.4, which is able to generate a model by means of a set of georeferenced images that are usually taken by a drone. Several flights were done with a \textit{Mavic 2 Pro} drone. The size of the modeled area was about 0.012 square kilometers with an average distance between samples of 0.56 cm. 
Moreover, to minimize the error localization of the model, some \textit{Ground Control Points} were taken in the experimental terrain by means of a RTK GNSS \textit{Emlid Reach M+} module, connected to the Andalusian Positioning Network (RAP) \cite{paez2017regional}, providing RTK accuracy of less than 10 cms. These points helped to rectify the model and remove a displacement in the UTM-Y axis of about 2.8 meters. Later on, Pix4Dmapper processed the resulting point cloud and produced a regular square grid of 0.1m resolution containing elevation data. It was later used to build up an hexagonal grid with 1m resolution for the anisotropic path planner.

\begin{figure}
    \centering
    \subfloat[Values of \textit{Steepness} $\alpha$.
	\label{fig-numericalSlope_umaTerrain}]%
    	{\includegraphics[width=0.48\columnwidth,align=c]{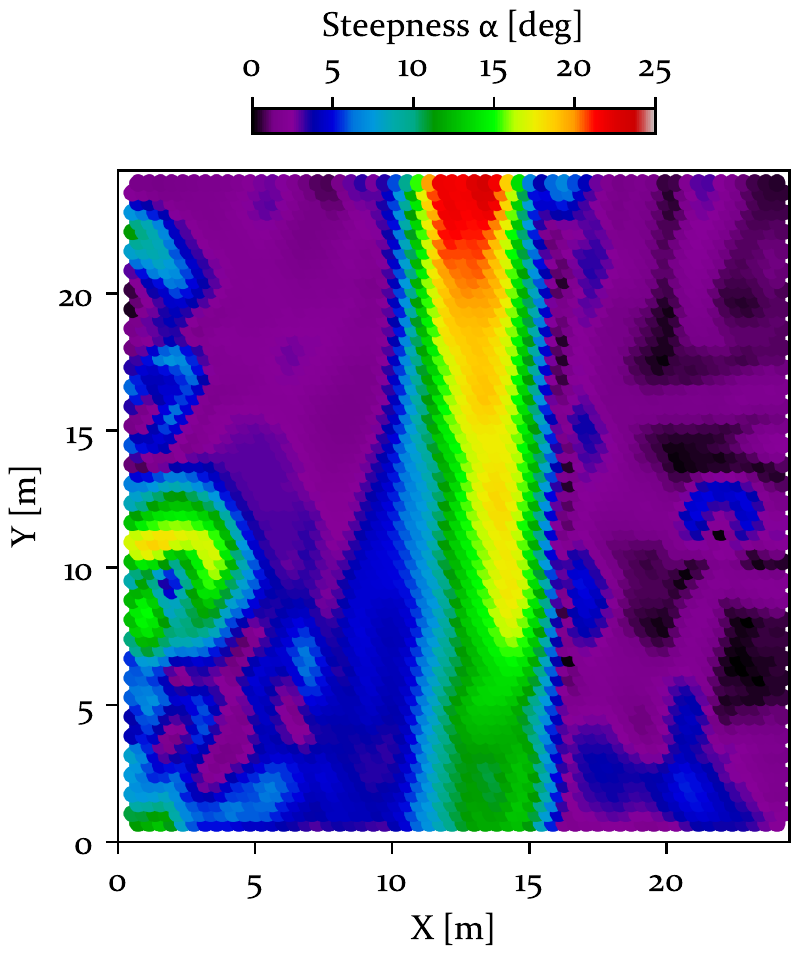}}
    \quad
	\subfloat[Values of aspect angle.
        \label{fig-numericalAspect_umaTerrain}]%
    	{\includegraphics[width=0.48\columnwidth,align=c]{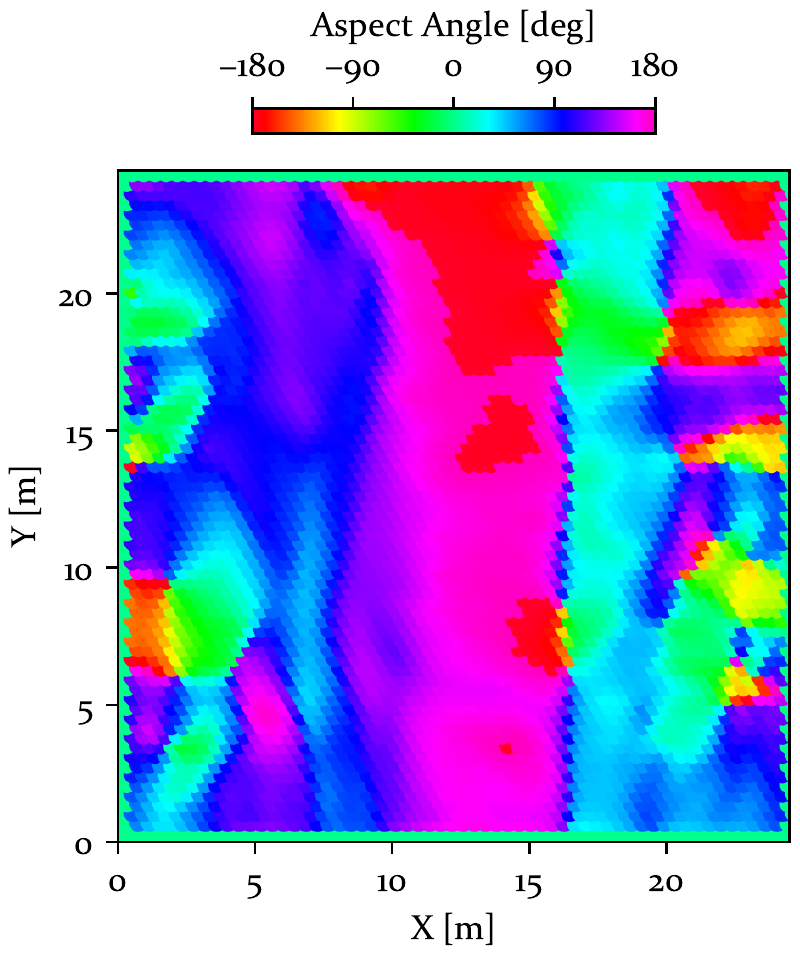}}
	\caption{Slope data describing the shape of the experimental terrain after smoothing the DEM with an average filter.}
    \label{fig-numericalMaps_umaTerrain}
\end{figure}

With regards the parameters to tune \textit{CAMIS}, they were customized according to the experimental mobile robotic platform \textit{Cuadriga}. Figures \ref{fig-cuadrigafront} and \ref{fig-cuadrigarear} introduce some images depicting different views of this robot. As can be observed, this robot is a four-wheeled skid-steering platform. It is commonly used for inspection tasks in rescue robotics field tests, and serves as a mobile testbed to study the performance of many mapping and navigation algorithms \cite{martinez2013navigability}.
It weights around 83 kilograms (a bit more depending on the instrumentation it carries). The wheels from each side are connected by a chain actuated by a motor that is controlled by a \textit{Roboteq} model AX1500 controller.
They receive power from a 36V battery pack with a capacity of 24Ah. Geometric information regarding the wheel arrangement can be found in previous research \cite{morales2010simplified}, where a preliminary version of this robot was used. In the current one, it is equipped with a chassis that increases its height to 0.81m. 

We made a series of preliminary traverses with this robot through various slopes located close to the one presented before. As result, we obtained the data shown in figure \ref{fig-cuadrigaData}. The current was measured using a \textit{ACS754} board with an \textit{Arduino AtMEGA} board, which is connected through an USB interface to the onboard computer (OBC). The orientation was obtained by interpolating the rover position on a processed DEM. As can be noted, the slip in this kind of terrain tends to zero, mainly because the soil is quite compact and the robot adheres well to the surface. From the readings of the current consumption, depicted in figure \ref{fig-adCuadrigaCostFunction}, a value of 0.45 was set to the \textit{resistance coefficient} $\rho$. As expected, with negative values of \textit{Pitch} $\theta$, i.e. when the robot ascends, the current consumption increases as consequence, while it decreases when the robot descends. The speed was around $0.5 m/s$ and, since the \textit{Cost} comes in terms of $As/m$, we defined $m = 2.43 As^2/m$, involving not only mass but also motor current torque, and, since this is modeled on Earth, $g = 9.8 m/s^2$. As the slip effects are negligible as demonstrated in figure \ref{fig-adCuadrigaSlipFunction}, we set $s_r = 0$ and $s_a = 0$.

\subsection{Roll Minimization}
\begin{figure}
    \centering
    \includegraphics[width=\columnwidth]{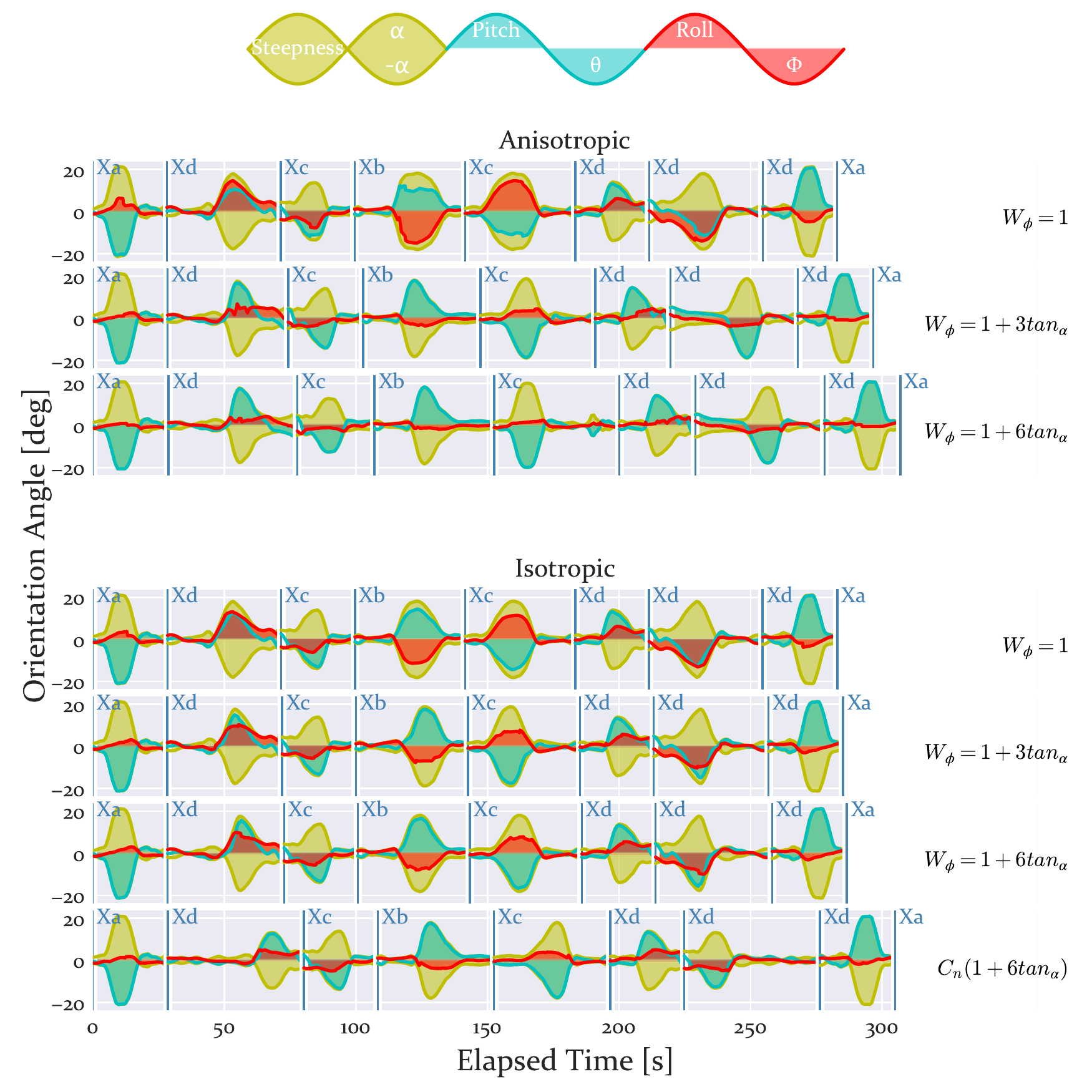}
    \caption{Orientation angles produced by each round-trip travel from the first three CAMIS model, their respective isotropic equivalent functions and the extra isotropic function introduced.}
    \label{fig-orientationResults}
\end{figure}
\begin{figure}
    \centering
    \includegraphics[width=\columnwidth]{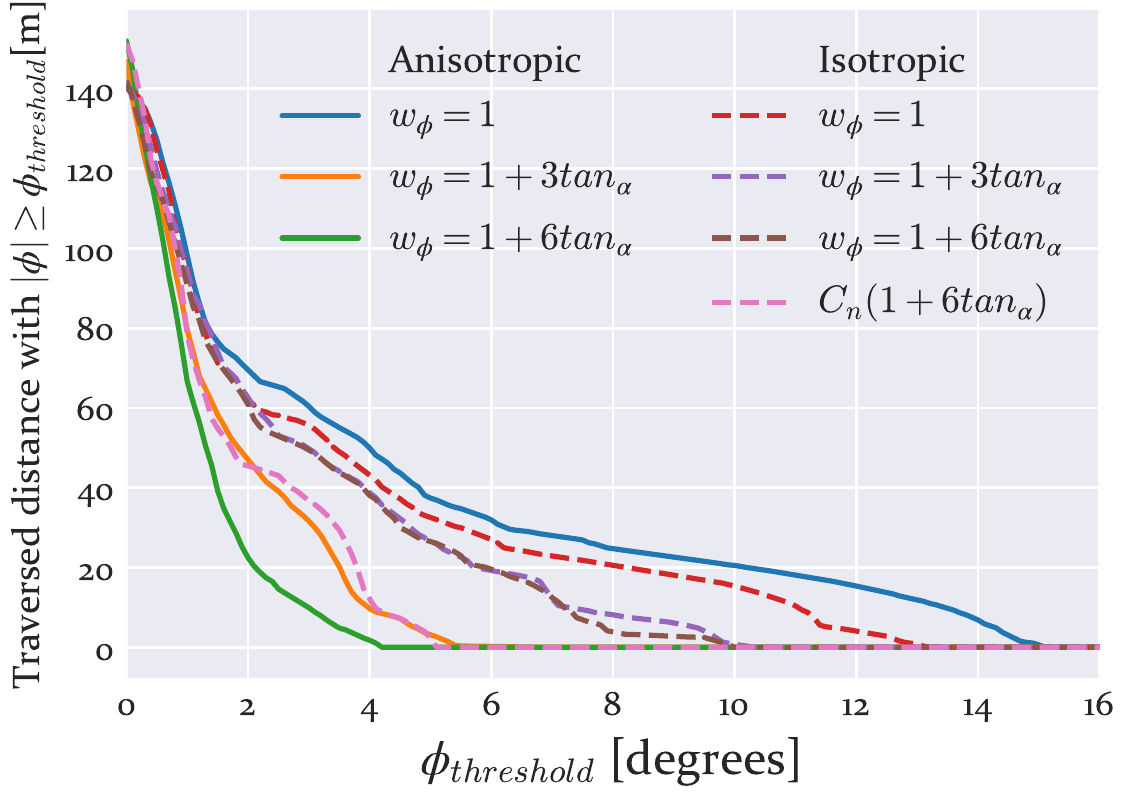}
    \caption{Traversed distance with an absolute value of \textit{Roll} $\phi$ higher than the threshold indicated in the horizontal axis.}
    \label{fig:rollPreservationResults}
\end{figure}
\begin{figure}
    \centering
    \includegraphics[width=1.0\columnwidth,align=c]{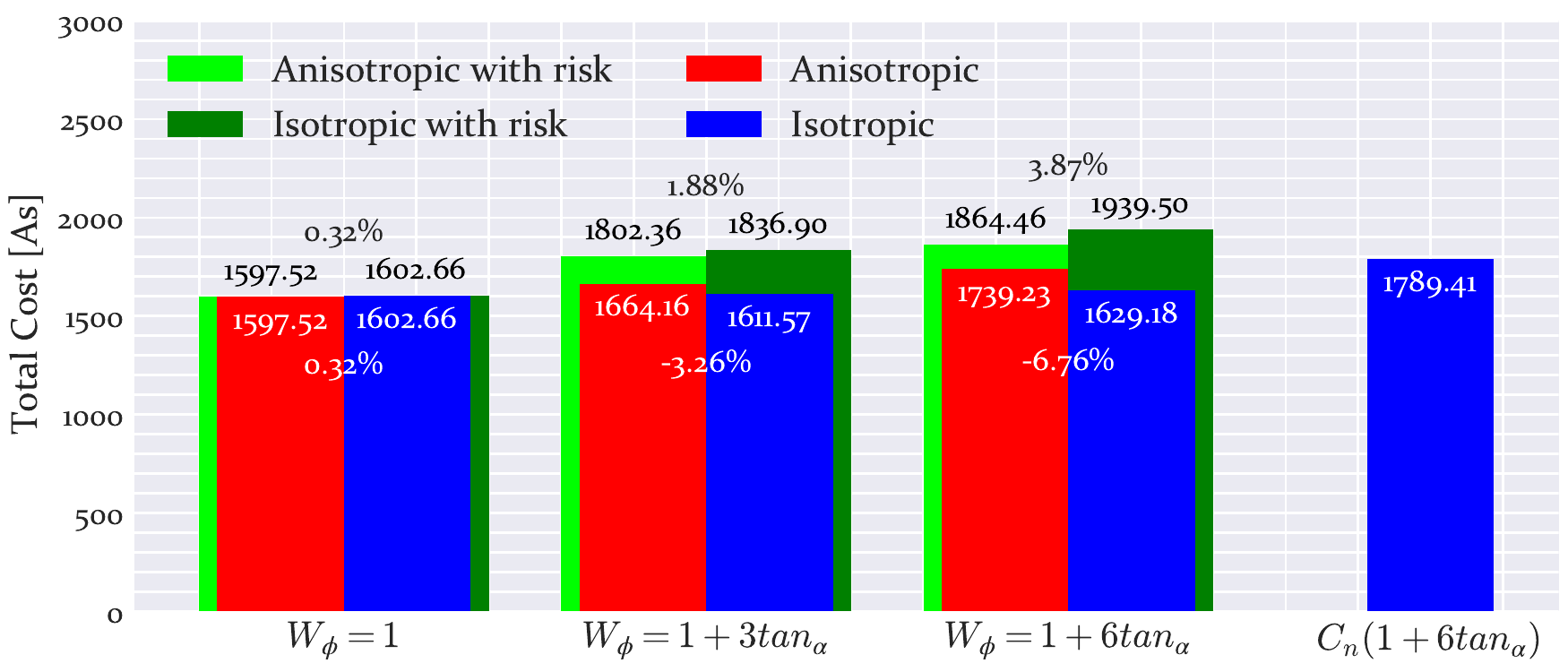}
    \caption{Estimated total cost (energy) for each configuration with and without the weight introduced by $w_{\phi}$. The green bars consider the \textit{Total Cost} corresponding not only to the energy consumption but also to the risk introduced by the \textit{Roll} weight function $w_{\phi}$.}
    \label{fig:costAndRisks}
\end{figure}

Then next simulation test focuses on studying the influence of the \textit{Roll} weight function $w_{\phi}$ in the angle of \textit{Roll} $\phi$ the robot experiences while climbing or descending through slopes. It is of high interest to check if $w_{\phi}$ can effectively modify the shape of the resulting path to make it go more parallel to the slope aspect vector $\gamma$, preventing it from going diagonally.
This is relevant, for example, to keep the lateral stability of the robot. As mentioned, the terrain surface for this test is part of the experimental field shown in figure \ref{fig-umaCuesta}. On top of the 3d model presented in figure \ref{fig-realCuesta3d} we indicate with the help of red arrows the location of four points of interest. These points are arranged by two, being each pair placed at each side of a slope. These positions are also marked in figure \ref{fig:roll_test_paths} and named $x_a$, $x_b$, $x_c$ and $x_d$, together with the portion of the terrain used. A deeper insight into the characteristics of this terrain can be found in figure \ref{fig-numericalMaps_umaTerrain}, which graphically presents the morphology of this terrain indicating the \textit{Steepness} and \textit{Aspect} angle of each location.

As can be observed in figure \ref{fig:roll_test_paths}, the resulting paths connecting the points of interest either ascend or descend through the slope that is located in the middle. There are two sequences to visit the points, each of them corresponding to the first two maps shown in this figure. In a similar manner to the previous test, we present on a third map those paths resulting from isotropic cost functions equivalent to the anisotropic ones. With regards defining the \textit{Roll} weight function, we make here use of three different expressions: $w_{\phi} = 1$, $w_{\phi} = 1 + 3 \tan_{\alpha}$ and $w_{\phi} = 1 + 6 \tan_{\alpha}$. The higher the rate at which the cost increases according to the \textit{Steepness} $\alpha$ the more parallel the paths are with respect to the \textit{Aspect}. This fact can be also checked in figure \ref{fig-orientationResults}, where the orientation angles corresponding to the traverse of each of the paths are plotted. In order to make this comparative more fair, we add another isotropic function. It is the isotropic equivalent to the \textit{CAMIS} case without \textit{Roll} weight ($w_{\phi} = 1$) but multiplied by a factor of $(1 + 6 \tan_{\alpha})$.

To better understand the benefit obtained from the anisotropic cost function, we introduce the plot shown in figure \ref{fig:rollPreservationResults}. This plot graphically represents the amount of meters in which the roll angle $\phi$ is above a certain threshold $\phi_{threshold}$. This threshold increases in the X-axis. As can be observed, the anisotropic cost function with $w_{\theta} = 1 + 6 \tan_{\alpha}$ greatly reduces this amount of distance above relatively high values of $\theta_{threshold}$, achieving the fact that no waypoints are above nearly 4 degrees of \textit{Roll}. On the other hand, the best isotropic case is the last one introduced, which gets closer to the anisotropic case where $w_{\theta} = 1 + 4 \tan_{\alpha}$.
As can be observed in figure \ref{fig:costAndRisks}, since the \textit{Total Cost} is a mix between \textit{Risk} introduced by $w_{\theta}$  and energy, the energy demanded by the anisotropic cases is higher than the isotropic counterparts. Nevertheless, the use of an isotropic function that focuses more on minimizing \textit{Roll}, $C_n(1 + 6 \tan_{\alpha)}$ in this case, demands more energy than the anisotropic functions.
In this way, we can conclude \textit{CAMIS} allows us to define a more refined trade-off between \textit{Roll} and energy consumption minimization.

\subsection{Field Tests}
\begin{figure*}
    \centering
    \includegraphics[width=1.0\textwidth,align=c]{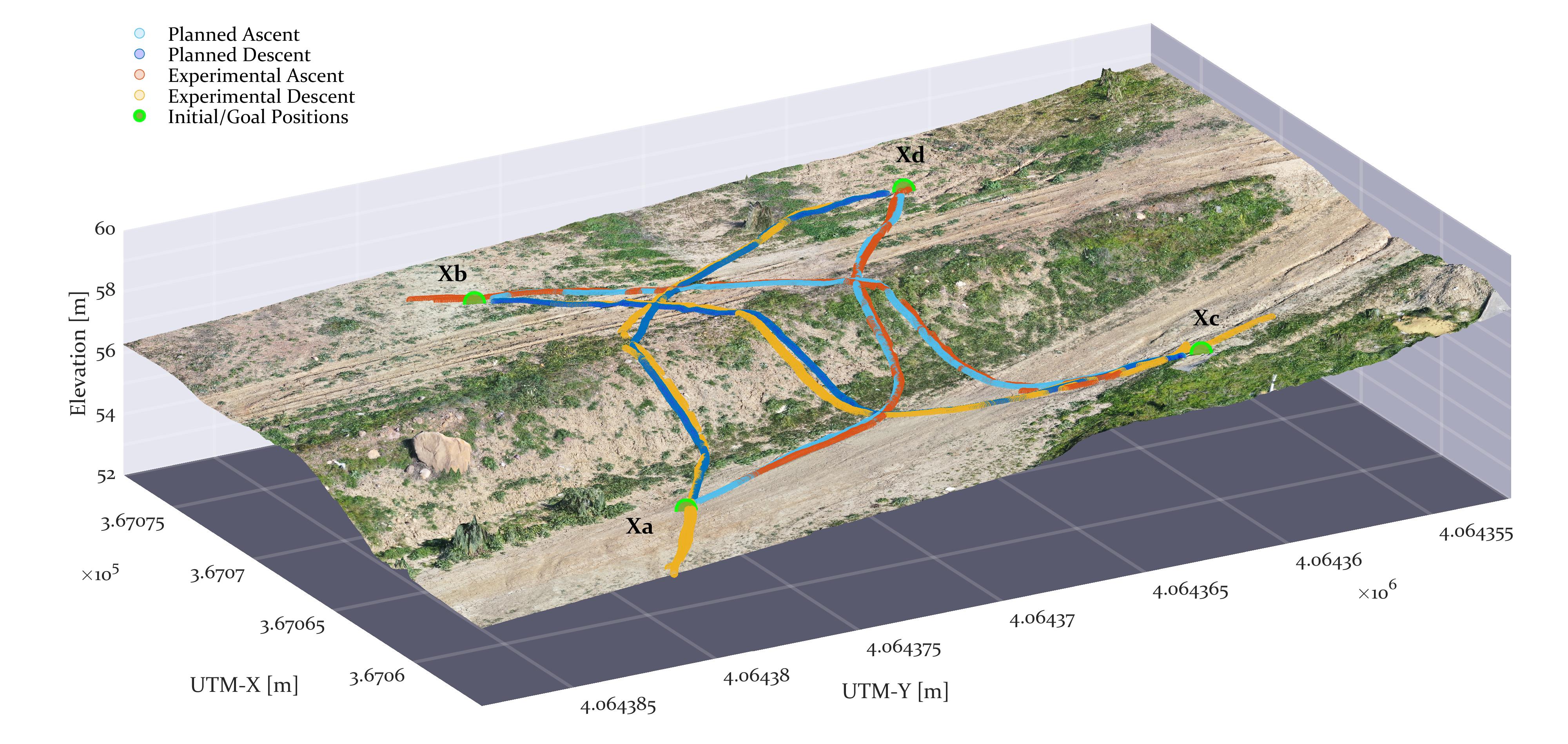}
    \caption{Paths traversing the slope with certain configurations of weight values.}
    \label{fig:cuadrigaPaths}
\end{figure*}
\begin{figure}
    \centering
    \includegraphics[width=\columnwidth,align=c]{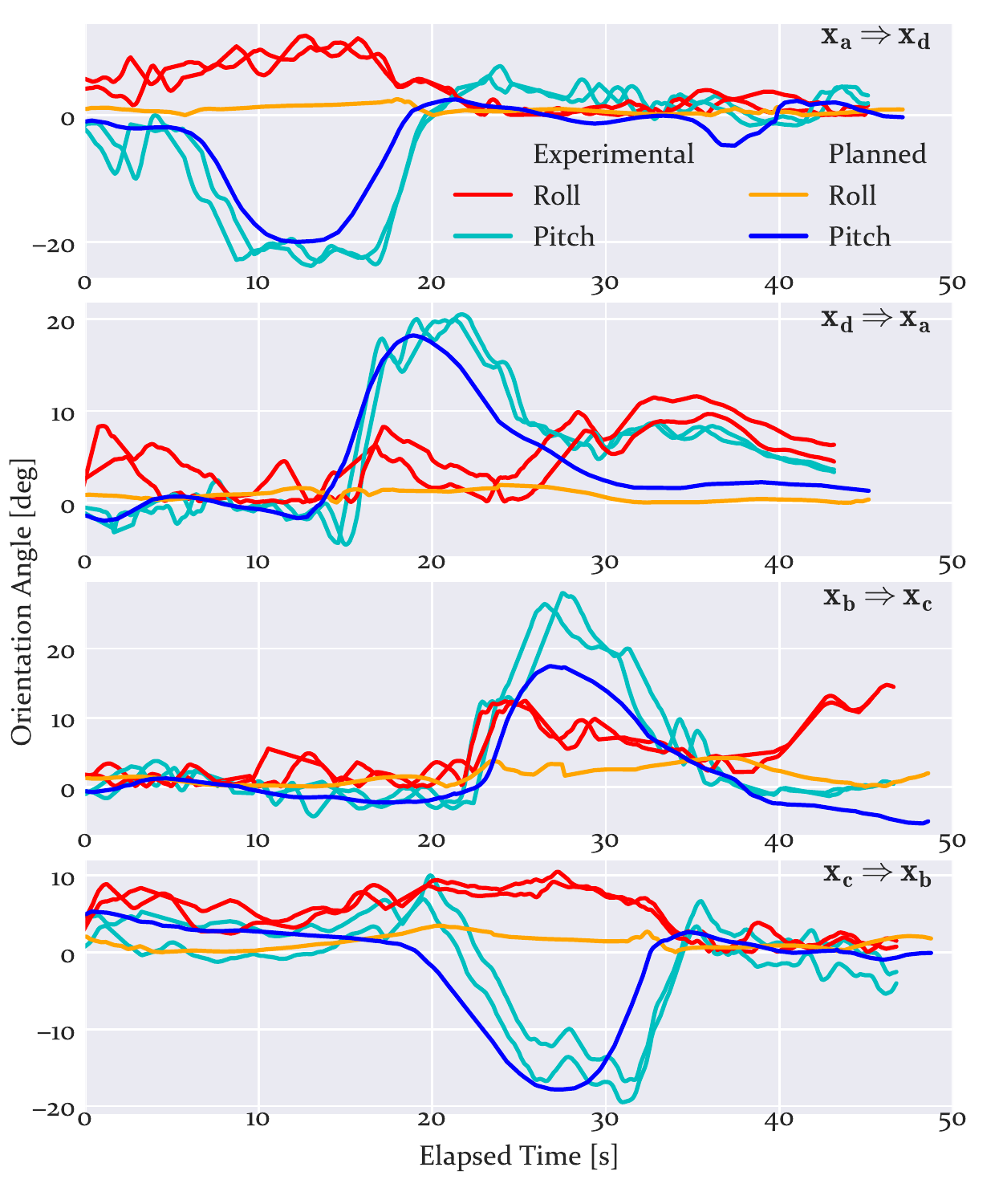}
    \caption{Comparative of orientation angles between simulated and experimental paths.}
    \label{fig:cuadrigaOrientation}
\end{figure}
\begin{figure}
    \centering
    \includegraphics[width=\columnwidth,align=c]{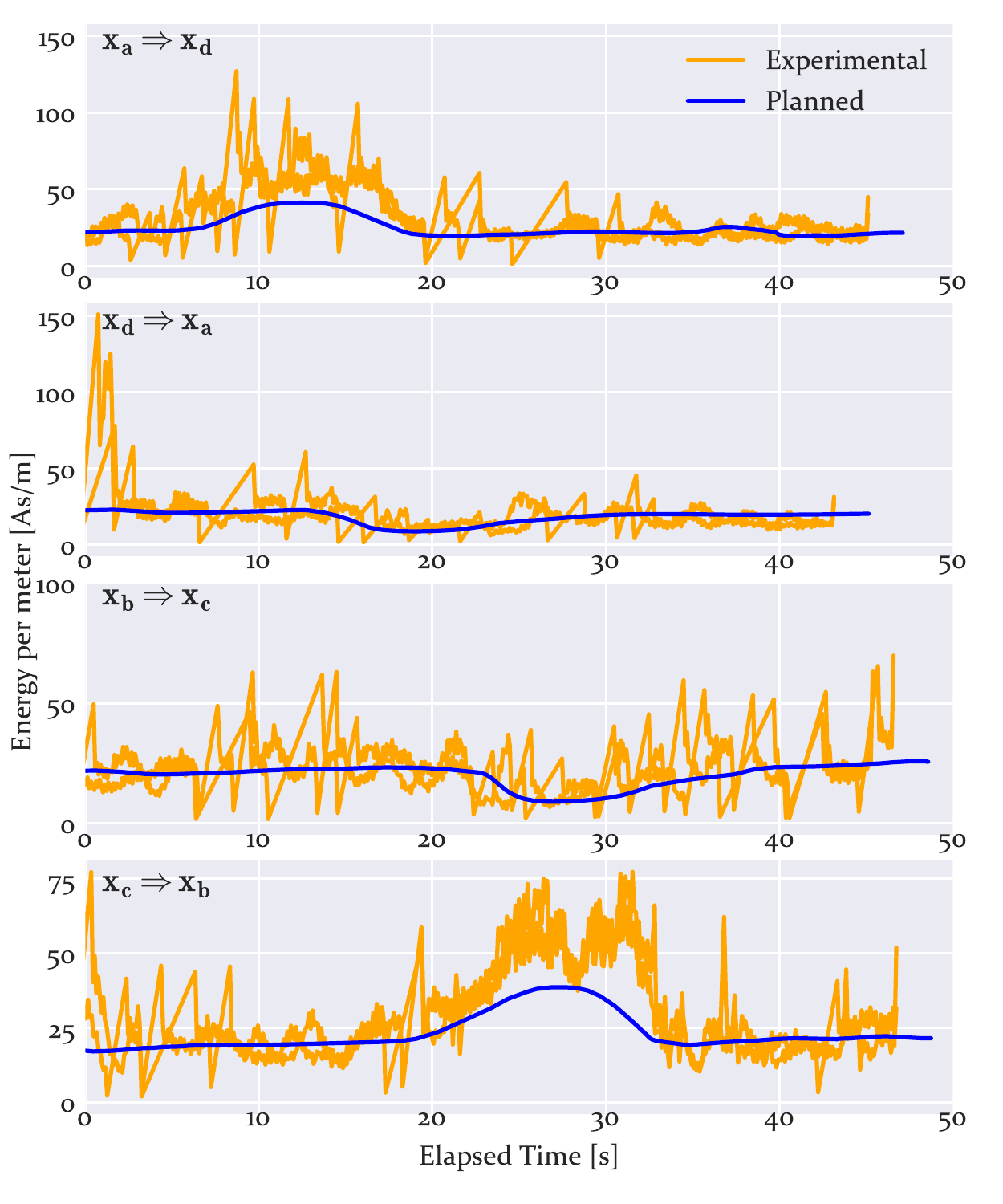}
    \caption{Comparative of the cost/speed between simulated and experimental paths.}
    \label{fig:cuadrigaConsumption}
\end{figure}
The last test is an experiment carried out with \textit{Cuadriga} on the real terrain. The main purpose of this test is to demonstrate how \textit{CAMIS} can be used in reality, analyzing the data collected from the drives and discussing the main issues found. The strategy we followed is to pick some paths from the last simulation test and make \textit{Cuadriga} follow them. The paths in question are those calculated by using the anisotropic cost function with $w_{\theta} = 1 + 6 \tan_{\alpha}$, connecting $x_a$ with $x_d$ and $x_b$ with $x_c$. Then, we make a comparative between the expected results from the planner and what the robot experienced during each drive, putting the focus on the orientation angles and the energy consumption. 

Figure \ref{fig:cuadrigaPaths} depicts the location of the points of interest $x_a$, $x_b$, $x_c$ and $x_d$, together with the waypoints that correspond to each planned path and the position samples of \textit{Cuadriga}. All this data is plotted on top of a 3d photo-realistic reconstruction of the terrain. As can be observed, the trajectory following made by \textit{Cuadriga} is quite accurate. The algorithm used is the \textit{Pure-Pursuit} \cite{campbell2007steering}, implemented in LabView and configured to use a look-ahead distance of 2m. The nominal speed, as in the preliminary drives to get the data for the model, is set to 0.5m/s. To solve the localization problem, the RTK GNSS \textit{Emlid Reach M+} module is installed on top of the vehicle. The orientation angles were measured using a \textit{MicroStrain 3DM-GX2} three-axis inertial measurement unit. 
A video showing how \textit{Cuadriga} drives through the slope following the paths is available online\footnote{\url{https://youtu.be/vJx_v2GRlSc}}. We carried out two executions with Cuadriga for each planned path, in order to reinforce the reliability of the results. For each execution the robot starts a bit further from the corresponding initial waypoint, to later stop at less than the look-ahead distance to the corresponding goal waypoint. With the data recorded from all drives, we made two comparatives with the outcome expected from the simulations. The first of them is presented in figure \ref{fig:cuadrigaOrientation}. It plots the \textit{Roll} and \textit{Pitch} angles that the robot experienced and those that were initially expected from the planner. We adjusted the elapsed time so in the case of the experimental results it begins when the robot is passing through the initial waypoint. As can be noted, the experimental drives finished a bit earlier due to the fact they stopped at a certain distance from the final waypoint. Although the experienced \textit{Pitch} angle seems quite similar to the planning, the \textit{Roll} deviates much more from the initial estimations. In the case where the trajectories go from $x_d$ to $x_a$ and from $x_b$ to $x_c$ it seems the IMU was not calibrated enough since in those cases \textit{Cuadriga} started very close to the initial waypoints, not leaving space to advance more meters in plain terrain unlike in the other cases. Besides, as can be noticed in figure \ref{fig:cuadrigaPaths} in the case where \textit{Cuadriga} went from $x_a$ to $x_d$ it did not exactly reproduce those pronounced turns existing in the planned paths, which made the angle of \textit{Roll} increase when smoothly turning on top of the slope. 
With regards the second comparative, figure \ref{fig:cuadrigaConsumption} plots the expected and experimented energy consumption per meter, which results from dividing the motor current consumption and the robot speed. In some cases this energy per meter took high values at the beginning as the rover started its movement and hence accelerated from the stopping position. The descent seems quite accurate while there is a certain error margin in the ascent. This may be because the conditions in which the preliminary drives were performed are slightly different from those in the testing drives, such as the presence of vegetation and differences in the pressure of the wheels.

\section{Conclusions}
\label{sec-conclusions}

In this paper we have presented a cost function model called \textit{CAMIS} aimed at performing anisotropic path planning on terrains containing slopes. By defining this model as the inverse polar function of a displaced ellipse, we make it compatible with the use of anisotropic PDE path planners like the bi-OUM. \textit{CAMIS} considers the energy consumption of the robot according to the direction it is facing with respect to any slope. Besides, it can also make a trade-off between minimizing this energy and preventing the robot from experiencing high values of \textit{Roll} angle. To better understand the use of \textit{CAMIS} we present in this paper two simulation tests and a field experiment involving a skid-steering robot.

The results from these tests have demonstrated in which situations this tool may be beneficial for making a robot optimally traverse scenarios with slopes. The results from the first simulation indicate that, in case of exclusively looking for minimizing the energy consumption, those vehicles affected more by slip effects will be most benefited from using \textit{CAMIS}. Moreover, according to the results from the second simulation, \textit{CAMIS} proves to be more versatile in making a polished trade-off to reduce both energy consumption and the experienced \textit{Roll} angle in contrast with isotropic cost functions. The field test served to check the fidelity of \textit{CAMIS} in a real situation, being an opportunity to discover many other influential elements that may be acknowledged in future versions, like the error in path tracking, which may introduce more \textit{Roll} angle, or certain terrain features like the roughness, which may make the \textit{Cost} increase.

Finally, we foresee the continuation of this work by studying the use of other PDE path planning methods such as Fast Sweeping, compatible with anisotropic cost functions. Besides, the use of an adaptive multi-resolution grid arises as a plausible solution to reduce the computational cost derived from the anisotropy and to prevent the apparition of local minima when it is relatively too high. Issues found like the discontinuities produced by the slip functions and the consideration of non-traversable areas are of interest for future work.

\ifCLASSOPTIONcaptionsoff
  \newpage
\fi

\bibliographystyle{IEEEtran}
\bibliography{IEEEabrv,bibliography}

\begin{IEEEbiography}[{\includegraphics[width=1in,height=1.25in,clip,keepaspectratio]{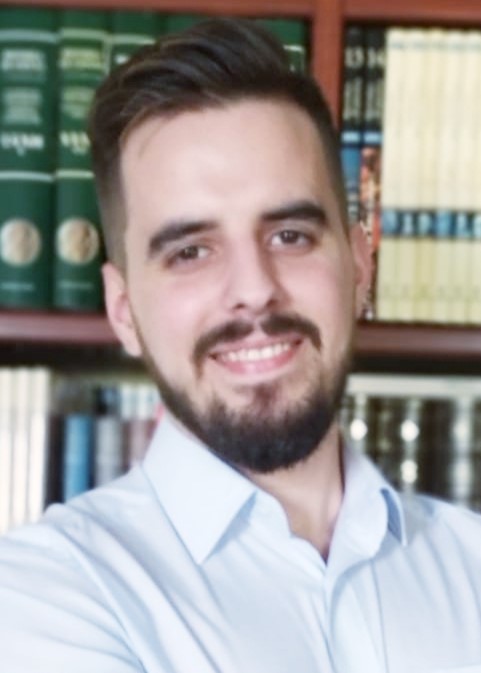}}]{J. Ricardo S\'anchez-Ib\'a\~nez}
received the M.Sc. degree in mechatronics from the University of Malaga in 2017, where he is currently coursing his Ph.D. studies. He performed three research stays between 2017 and 2019 in the Automation and Robotics Section at ESA-ESTEC, with a total duration of one year. He worked in the FIRST-ROB project, funded by the spanish government, and in the H2020 ADE project, funded by the European Commission. In the context of the latter he made a one-month visit to the DFKI Robotics and Innovation Center in Bremen, Germany. His research interests include space robotics and autonomous navigation.
\end{IEEEbiography}

\begin{IEEEbiography}[{\includegraphics[width=1in,height=1.25in,clip,keepaspectratio]{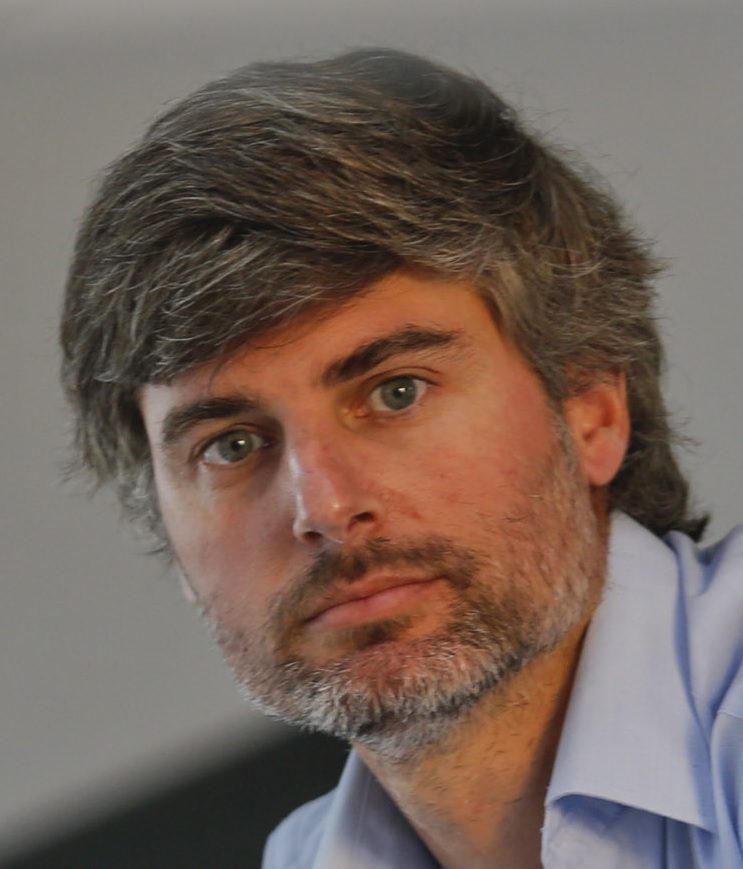}}]{Carlos P\'erez-Del-Pulgar}
 received M.S. degree and PhD in Computer Science at the University of Málaga (UMA). In 2004, he was given a permanent position as research support staff at the University of Malaga until 2017. Also, since 2010 he was part-time assistant lecturer in the Electrical Engineering Faculty where he was responsible of different subjects related to automation and robotics. Currently he is associate lecturer at the same University. In 2014 he performed a research stay in the Automation and Robotics section at ESA-ESTEC. His research interests include machine learning, surgical robotics, space robotics and autonomous astronomy. He has more than 60 publications in these topics and he has been involved in more than 15 different Spanish and European projects.
\end{IEEEbiography}

\begin{IEEEbiography}[{\includegraphics[width=1in,height=1.25in,clip,keepaspectratio]{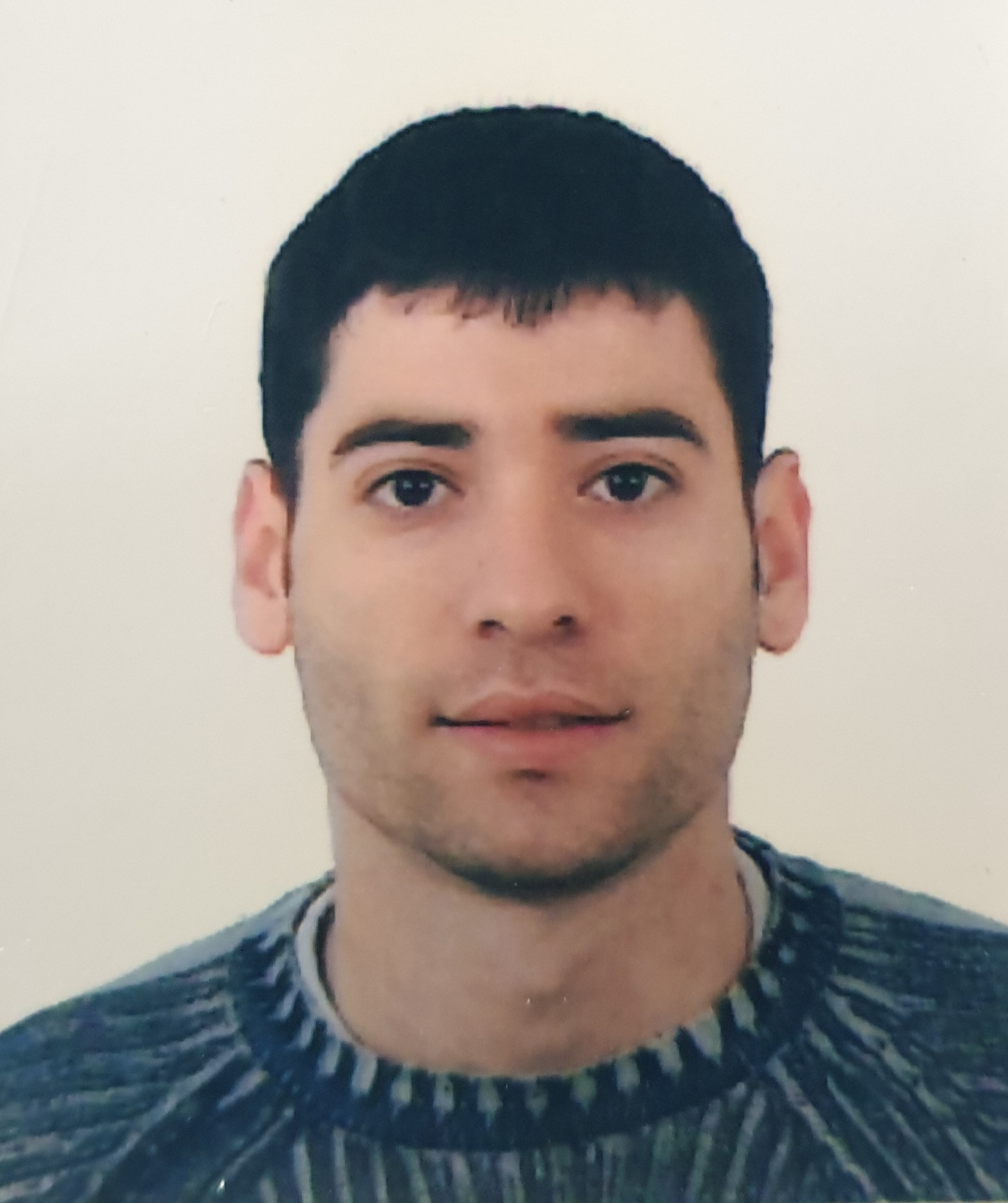}}]{Javier Ser\'on} received the M. Eng. degree in Electrical and Electronic Engineering and the Ph.D. degree from the University of Malaga in 2000 and 2012, respectively. In 2000, he joined the Engineering Systems and Automation Research Group as a Researcher. His research interests include process automation, special manipulators, telerobotics, mobile robots and autonomous vehicles. He participated in the construction of a surgical assistant, a special manipulator (Goniophotometer) and several mobile robots. 
\end{IEEEbiography}

\begin{IEEEbiography}[{\includegraphics[width=1in,height=1.25in,clip,keepaspectratio]{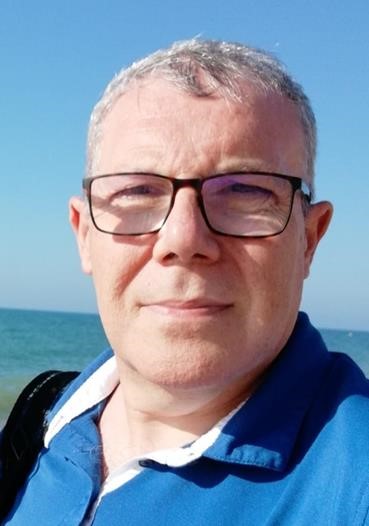}}]{Alfonso Garc\'ia-Cerezo} was born in 1959 in Sigüenza (Spain). He received the Industrial Electrical Engineer and the Doctoral Engineer Degree from the Escuela Tecnica Superior de Ingenieros Industriales of Vigo in 1983 and 1987, respectively.
From 1983 to 1988 he was Associate Professor in the Department of Electrical Engineering, Computers and Systems at the University of Santiago de Compostela. From 1988 to 1991 he was Assistant Professor at the same university.
Since 1992 he has been a Professor of System Engineering and Automation at the University of Malaga, Spain. Since 1993 to 2004 he has been Director of the “Escuela Tecnica Superior de Ingenieros Industriales de Malaga”. He is now the Head of the Department, and the responsible of the “Instituto de Automática Avanzada y Robótica de Andalucía” in Málaga.
He has authored or co-authored about 150 journal articles, conference papers, book chapters and technical reports. His current research included mobile robots and autonomous vehicles, surgical robotics, mechatronics and intelligent control.
Besides, is the responsible of more than 19 research projects during the last 10 years.
He is also a member of several international and national scientific and technical societies like IEEE, EUROBOTICS, IFAC,  AER-ATP, CEA and SEIDROB.
\end{IEEEbiography}

\end{document}